\documentclass[10pt,twocolumn,letterpaper]{article}

\usepackage{cvpr}              % To produce the CAMERA-READY version

\usepackage{graphicx}
\usepackage{amsmath}
\usepackage{amssymb}
\usepackage{booktabs}
\usepackage[dvipsnames]{xcolor}
\usepackage{makecell}
\usepackage{multirow}
\usepackage{hhline}
\usepackage{float}
\usepackage{algorithm}
\usepackage{comment}
\usepackage[noEnd=True]{algpseudocodex}

\usepackage[dvipsnames]{xcolor}

\usepackage{graphicx}
\usepackage{amsmath}
\usepackage{amssymb}
\usepackage{booktabs}
\usepackage[dvipsnames]{xcolor}
\definecolor{kaimingblue}{HTML}{0071bc}
\definecolor{universityred}{HTML}{990000}

\definecolor{cvprblue}{rgb}{0.21,0.49,0.74}
\definecolor{abbrcolor}{HTML}{990000}
\definecolor{figureorange}{HTML}{D79B00}
\definecolor{ourred}{HTML}{990000}
\definecolor{ourpurple}{HTML}{9673A6}
\definecolor{ourorange}{HTML}{D79B00}
\definecolor{ourblue}{HTML}{6C8EBF}
\usepackage[pagebackref,breaklinks,colorlinks,linkcolor=ourred,citecolor=cvprblue]{hyperref}
\usepackage[hang,flushmargin]{footmisc}

\DeclareMathOperator*{\argmax}{argmax}
\newcommand{\ourframework}{UMO\xspace}
\newcommand{\genmodel}{$\mathcal{G}_\phi$\xspace}
\newcommand{\targetmodel}{$f_\theta$\xspace}
\newcommand{\clipimage}{$\mathcal{E}_I$\xspace}
\newcommand{\cliptext}{$\mathcal{E}_T$\xspace}
\newcommand{\clipmodel}{$\mathcal{C}$\xspace}
\newcommand{\originalimg}{$x_i$\xspace}
\newcommand{\counterfactualimg}{$\hat{x}_i$\xspace}

\newcommand{\figa}{\textcolor{ourred}{(a)}\xspace}
\newcommand{\figb}{\textcolor{ourpurple}{(b)}\xspace}
\newcommand{\figc}{\textcolor{ourorange}{(c)}\xspace}

\newcolumntype{C}[1]{>{\centering\arraybackslash}p{#1}}

\def\paperID{4281} % *** Enter the Paper ID here
\def\confName{CVPR}
\def\confYear{2024}

\title{Unsupervised Model Diagnosis}

\author{Yinong Oliver Wang\textsuperscript{1} \qquad Eileen Li\textsuperscript{1} \qquad Jinqi Luo\textsuperscript{2} \qquad Zhaoning Wang\textsuperscript{3} \qquad Fernando De la Torre\textsuperscript{1}\\
Carnegie Mellon University\textsuperscript{1} \qquad 
University of Pennsylvania\textsuperscript{2} \qquad 
University of Central Florida\textsuperscript{3}\\
{\tt\small \{yinongwa, chenyil, ftorre\}@cs.cmu.edu \qquad 
jinqiluo@upenn.edu \qquad
zhaoning@ucf.edu }
}

\begin{document}
\maketitle

\let\thefootnote\relax
\footnote{Preprint. Under review.}
\footnote{Code will be available soon.}

\begin{abstract}
Ensuring model explainability and robustness is essential for reliable deployment of deep vision systems. Current methods for evaluating robustness rely on collecting and annotating extensive test sets. While this is common practice, the process is labor-intensive and expensive with no guarantee of sufficient coverage across attributes of interest. Recently, model diagnosis frameworks have emerged leveraging user inputs (e.g., text) to assess the vulnerability of the model. However, such dependence on human can introduce bias and limitation given the domain knowledge of particular users. 

This paper proposes \textcolor{abbrcolor}{\textbf{U}}nsupervised \textcolor{abbrcolor}{\textbf{Mo}}del Diagnosis (\ourframework), that leverages generative models to produce semantic counterfactual explanations without any user guidance. Given a differentiable computer vision model (i.e., the target model), \ourframework optimizes for the most counterfactual directions in a generative latent space. Our approach identifies and visualizes changes in semantics, and then matches these changes to attributes from wide-ranging text sources, such as dictionaries or language models. We validate the framework on multiple vision tasks (e.g., classification, segmentation, keypoint detection). Extensive experiments show that our unsupervised discovery of semantic directions can correctly highlight spurious correlations and visualize the failure mode of target models, \textbf{without any human intervention}.

\end{abstract}    
\vspace{-1mm}
\section{Introduction}
\label{sec:introduction}
\vspace{-1mm}
\begin{figure}[!t]
\centering
  \includegraphics[width=\linewidth]{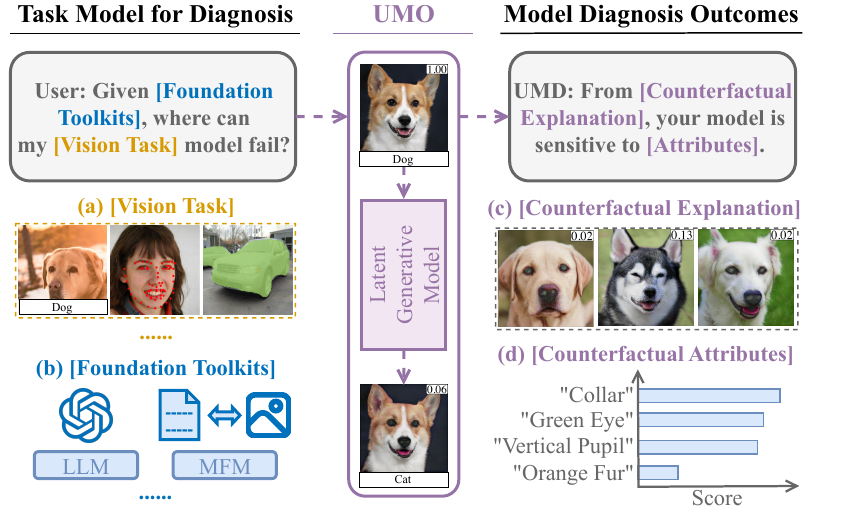}
\caption{\textbf{Overview.} Given a \textcolor{ourorange}{(a)} computer vision model (\eg, classifier, key-point detector, segmentation model), how can we understand the model vulnerabilities without requiring user input nor test sets? Our proposed framework \ourframework leverages \textcolor{ourblue}{(b)} foundation toolkits (\eg, large language models (LLMs) and multi-modal foundation models (MFMs)) to perform \textbf{unsupervised} model diagnosis. \ourframework not only outputs \textcolor{ourpurple}{(c)} counterfactual visual explanations but also \textcolor{ourpurple}{(d)} top-matched counterfactual attributes.
}
\label{fig:first_fig}
\vspace{-4mm}
\end{figure}

Contemporary methods for assessing computer vision algorithms primarily rely on the evaluation of manually labeled test sets. However, relying solely on metric analysis of test sets does not ensure the robustness or fairness of algorithms in real-world scenarios~\cite{duan2020adversarial}. This limitation arises from several factors. Firstly, while this approach is effective at gauging performance under known test conditions, it does not proactively address unforeseen model failures. Secondly, it is often infeasible to gather test sets that encompass all potential scenarios or relevant attributes of interest. Lastly, the process of constructing test sets is typically resource-intensive, time-consuming, and susceptible to errors. To address such issues, this paper leverages large pre-trained models (LPMs), trained on extensive datasets comprising millions of samples, as a means to assess potential shortcomings in computer vision models. The question that we try to address is: \textit{Can these LPMs be applied to evaluate various computer vision tasks (e.g., segmentation) by uncovering possible failure modes and limitations in a completely \textbf{unsupervised} manner? }

An emergent approach to discover model failure modes without exhaustive test sets makes use of counterfactual explanations~\cite{Mothilal_2020, pmlr-v97-goyal19a}. Along this line of research, model defects are studied through challenging images that lead to model failure modes. However, earlier efforts deceive the target models via pixel-level adversarial perturbations~\cite{goodfellow2014fgsm,madry2018towards}, which are not informative and do not explain model failures. A few works produce adversarial examples by semantically perturbing base images along semantics~\cite{2019SemanticAE,2020semanticadv}, but they focus on effective attacks rather than model evaluation. To gain insights into a model's flaws, we need counterfactual explanations that reveal the semantic differences that lead to model failures. Recently, \cite{luo2023zeroshot} proposed a zero-shot method to analyze model sensitivity to attributes via counterfactual explanation. Nonetheless, this method still requires domain knowledge from the user about potential attributes of interest, which can introduce biases and limit its outcome to the domain knowledge of particular users.

To address the issues mentioned above, we introduce Unsupervised Model Diagnosis (\ourframework) to discover the model failures and perform open-vocabulary model diagnosis without any user input or domain knowledge. \cref{fig:first_fig} illustrates the 
main idea of our work. \ourframework comprises two main input components:
\textcolor{ourorange}{(a)} a target model (for vision tasks) provided by the user; in this paper, we address classification, segmentation, and key-point detection,
\textcolor{ourblue}{(b)} a collection of foundation toolkits, \ie LPMs, for counterfactual image generation as well as language models for semantic analysis.
The output of \ourframework is a diagnostic report that includes \textcolor{ourpurple}{(c)} a set of visual counterfactual explanations and \textcolor{ourpurple}{(d)} the corresponding list of counterfactual image attributes with their associated semantic similarity scores, see ~\cref{fig:first_fig} right. 

The resulting diagnosis offers insights into the specific visual attributes to which the model is vulnerable. This information can guide actions such as collecting additional data for these attributes or adjusting their weights in the training process. Our complete pipeline operates in an unsupervised manner, eliminating the requirement for data collection and label annotation. 

To summarize, our proposed work brings two main advancements when compared to previous efforts:

\begin{itemize}
   \item We propose an unsupervised framework for model diagnosis, named \ourframework, that bypasses the tedious and expensive requirement of human annotation and user inputs.
   \item \ourframework utilizes the parametric knowledge from foundation models to ensure accurate analysis of model vulnerabilities. The framework does not require a manually-defined list of attributes to generate counterfactual examples.
   % \item \ourframework paves the way for enhancing models, providing key insights that guide further improvements.
\end{itemize}
% \vspace{-1mm}
\section{Related Work}
\label{sec:related_work}
\vspace{-1mm}

This section discusses past work on vision model diagnosis and visual manipulation through generative latent space.

\subsection{Latent Generative Models}

Generative models, especially StyleGAN \cite{2019stylegan,Karras2020ada,styleganxl} and Diffusion Models \cite{ddpm,ddim,rombach2021highresolution}, have semantic latent spaces that are differentiable, and can be used to edit image attributes \cite{sefa, ganspace, interfacegan}. StyleSpace~\cite{stylespace} found a more disentangled latent space by manipulating the modulation weights (\ie, style codes) in StyleGAN affine transformation. One step further, StyleCLIP~\cite{2021StyleCLIP} guided the generation sampled from StyleSpace by minimizing the CLIP~\cite{clip} loss between the sampled image and the user text prompt. Similarly, DiffusionCLIP~\cite{kim2022diffusionclip} applies image modification using text pairs via CLIP directional loss. Note that both StyleCLIP and DiffusionCLIP aim to apply targeted edits given by the user, whereas UMO aims to learn critical edits in an unsupervised fashion.
Recent advances in large language models enable informative supervision of the generation process. \cite{zhang2023controllable, lian2023llmgrounded} use language models to generate multimodal conditions for enhanced compositionality and reasoning of diffusion models. In the line of controlling diffusion latent space, ControlNet \cite{zhang2023adding} learned a task-specific network to impose constraints on Stable Diffusion \cite{rombach2021highresolution}. Asymmetric reverse process (Asyrp) \cite{asyrp} discovered that the bottleneck layer of the U-Net in the diffusion model encodes meaningful semantics; hence, the authors proposed to learn a unified semantic edit direction applicable across all images in this bottleneck latent space. Our work adopts the similar concept of optimizing globally effective counterfactual directions for the target model. By manipulating the semantics and analyzing the learned direction, we can gain valuable insights into model failure visualization and bias identification.
\begin{figure*}[!t]
    \centering
    \vspace{-4mm}
    \includegraphics[width=0.95\linewidth]{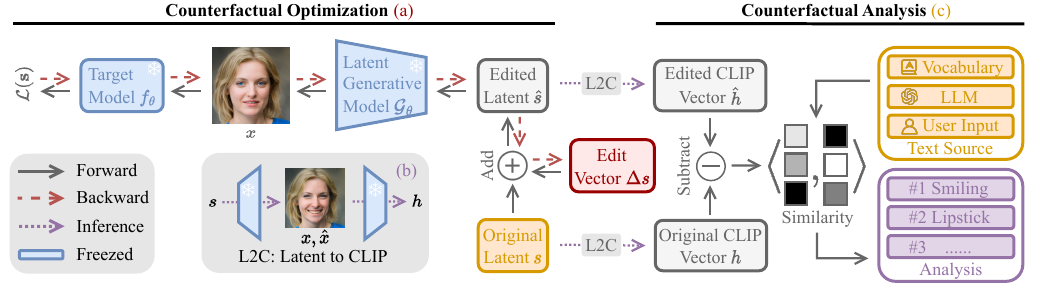}
    \vspace{-1mm}
    \caption{\textbf{The \ourframework framework.} Black solid lines denote forward passes; red dashed lines denote backpropagation; and purple dotted lines denote the inference of analysis. \figa We first optimize an edit direction $\Delta s$ in the latent space of generative models that yields counterfactual images of the target model. \figb After the optimization converges, we generate the original and edited images $x$ and $\hat{x}$ and map them to the CLIP embedding space with the L2C block (in \cref{sec:counterfactual_analysis}). \figc Then we analyze and report the diagnosis of counterfactual attributes by matching the image embedding differences $\hat{h}-h$ with attribute candidates. 
    }
    \label{fig:method}
    \vspace{-2mm}
\end{figure*}

\vspace{-0.5mm}
\subsection{Diagnosis of Computer Vision Models}
\vspace{-0.5mm}
Model diagnostics \cite{davision1992regression} originally referred to the validity assessment of a regression model, including assumption exploration and structural examination. Diverging from typical adversarial perturbations~\cite{shukla2023generating,wang2023generating}, whose sole objective was to make the model fail, recent years have witnessed the trend of the vision community broadly adopting the term \textit{diagnose} \cite{zhang2023diagnosing,luo2023zeroshot,vendrow2023dataset,prabhu2023lance} for understanding and evaluating the failure of deep vision models, particularly focusing on attribution bias, adversarial robustness, and decision interpretability to identify potential flaws. To search model failure cases, \cite{goodfellow2014fgsm} first proposed pixel-space perturbations by signed gradient ascent to generate adversarial examples. \cite{madry2018towards} further advanced the philosophy by multi-step gradient projection, and \cite{2018GANAE} proposed to synthesize the pixel-space perturbation by generative model. 

However, \cite{2019SemanticAE,2020semanticadv} claimed that such adversaries lack visual interpretability and proposed attacking the model by optimizing along fixed semantic axes of generative models. Similarly, in the literature of counterfactual explanation \cite{3}, methods \cite{4,5,6,8,9,10,Mothilal_2020} commonly focus on generating semantic attacks on a per-image basis, which overlook the global model-centric vulnerability diagnosis. Despite the effective instance-level counterfactual generation, such failure-driven attacks can be less informative for diagnosing model vulnerabilities (\eg, altering the perceived gender to fool a gender classifier). Hence, when directly applying these attack-oriented counterfactual explanations for model diagnosis, they usually require additional human interpretation to summarize individual failures. Besides being often designed for specific single task \cite{perez2023towards,5,9}, previous methods \cite{7,8,10} also require fine-tuning of the generative pipeline. Hence, we adopt a diagnosis-driven, task-agnostic, and resource-efficient counterfactual synthesis pipeline desirable for diagnosing instead of simply attacking models.

To emphasize the explainability requirement of model diagnosis and produce human-understandable outcomes, recent works visualized model failures by optimizing an attribute hyperplane \cite{li-2021-discover}, identifying error subgroups from cross-modality gaps \cite{zhang2023diagnosing, eyuboglu2022domino}, recognizing sensitive style coordinates \cite{Lang_2021_ICCV}, searching semantic variations by unconditional generative models \cite{li2022attractive,jeanneret2023adversarial}, or fine-tuning language model to perturb prompts for text-condition generations \cite{prabhu2023lance}. Nevertheless, these approaches require either manually annotating the discovered failure and the collected test set, or training a model-specific explanation space for every new target model. Our approach addresses these shortcomings by performing diagnosis in an unsupervised manner with the help of foundation toolkits. More recently, ZOOM \cite{luo2023zeroshot} proposed to analyze provided attributes in a zero-shot manner. Our method distinguishes itself from ZOOM in that: (1) \ourframework achieves automatic discovery instead of focusing on analyzing given attributes, (2) \ourframework is more exploratory in counterfactual edits while ZOOM can only analyze in restricted attribute directions, (3) \ourframework requires no prior knowledge from the user and hence circumvents potential human biases.

\vspace{-1mm}
\section{Method}
\label{sec:method}
\vspace{-1mm}

%Observing the lack of an easily accessible and generally applicable diagnosis method for computer vision models, we leverage recent advances in 
%LPMs to propose \ourframework, an end-to-end unsupervised model diagnosis model. 

Given a target model \targetmodel, our pipeline consists of two stages as shown in \cref{fig:method}. In the first stage, a latent generative model, denoted as \genmodel, (\eg, a Diffusion model or GAN) is used to discover counterfactual modifications, denoted as $\text{UMO}(f_\theta, \mathcal{G}_\phi)$, by directly optimizing the latent edit directions that can mislead the prediction of \targetmodel, shown in \cref{fig:method}(a). In the second stage, after the counterfactual optimization converges, we generate pairs of image embeddings (original, counterfactual) in \cref{fig:method}(b). With these pairs, we are able to interpret and analyze the counterfactual attributes of the target model \targetmodel by computing semantic similarity scores with attribute candidates, as illustrated in \cref{fig:method}(c).

\subsection{Counterfactual Optimization}
\label{sec:counterfactual_optimization}

Given a target model \targetmodel to diagnose, we first focus on discovering cases that lead to incorrect model predictions. To capture the failure modes, an effective solution is to generate counterfactual examples of the target model. Hence, our first step aims to learn latent edits representing semantic edit directions. When injecting such latent modification to the generative model \genmodel, the generated images are observed to have meaningful semantic changes~\cite{li2021exploring} but are challenging to the target model. We denote each pair of the original image \originalimg and its edited adversarial counterpart \counterfactualimg as a counterfactual pair. This section shows that \ourframework can discover and generate critical semantic counterfactual pairs for a given target model \targetmodel using various generative models \genmodel (\eg, StyleGAN and Diffusion Model).

Since the StyleSpace $\mathcal{S}$ is shown to be effective for semantic manipulation \cite{stylespace,2021StyleCLIP}, we choose to inject counterfactual edits in this latent space. We first initialize our latent edit vector $\Delta s$ in the space $\mathcal{S}$ as Gaussian noise. Then we sample $N$ style vectors $\{s_i\}_{i=1}^N \sim \mathcal{S}$ and generate $N$ corresponding images $x_i = \mathcal{G}_\phi(s_i)$. Note that optionally real images $x_i$ can also be used and $s_i$ is then obtained through GAN inversion\cite{ganinversion}; but for the rest of the paper, we use synthetic images $x_i$ to leverage free exploration in generative latent space for more diverse counterfactual explanations. For each original image \originalimg, we inject the edit vector $\Delta s$ to the latent space $\mathcal{S}$ and obtain the edited image $\hat{x}_i = \mathcal{G}_\phi(s_i + \Delta s)$. With the original and edited image based on the same initial latent vector, we compute the following loss $\mathcal{L}$: 
\begin{equation}
\label{eq:total_loss}
    \mathcal{L}(\Delta s) = 
    \alpha \mathcal{L}_{\text{target}} +
    \beta \mathcal{L}_{\text{CLIP}} +
    \gamma \mathcal{L}_{\text{SSIM}} +
    \mathcal{L}_{\text{reg}}.
\end{equation}
% \begin{equation}
% \label{eq:total_loss}
% \mathcal{L}(\Delta s) = 
% \alpha \mathcal{L}_{\text{target}}(\hat{x}_i) +
% \beta \mathcal{L}_{\text{CLIP}}(\hat{x}_i) +
% \gamma \mathcal{L}_{\text{SSIM}}(\hat{x}_i) + 
% \mathcal{L}_{\text{reg}}(\Delta s)
% \end{equation}
To optimize w.r.t. $\Delta s$, we back-propagate the loss as shown in the backward process in \cref{fig:method}(a).

The first component of the loss, $\mathcal{L}_{\text{target}}$, 
is the adversarial loss that ensures the learned $\Delta s$ can edit the original image to cause target model failure. This loss measures the distance between the model prediction on the edited image $f_\theta(\hat{x}_i)$ and the opposite of the original model prediction $f_\theta(x_i)$. It is task-dependent: when diagnosing a binary classification task, we want to minimize the cross-entropy loss between $f_\theta(\hat{x}_i)$ and $\hat{p}_{\text{cls}} = 1-f_\theta(x_i)$ such that the edits effectively mislead the classifier toward the opposite class; when diagnosing a keypoint detector or a segmentation model, we want the (perturbed) incorrect model predictions to be close to a randomized or targeted pseudo label (details in \cref{sec:other_tasks}), denoted as $\hat{p}_{\text{kdet}}$ and $\hat{p}_{\text{seg}}$ respectively:
\begin{align}
    &\text{(binary classifier) }\mathcal{L}_{\text{target}}({\hat{x}}_i) = {L}_{\text{CE}}(f_{\theta}({\hat{x}}_i), \hat{p}_{\text{cls}}),\\
    &\text{(keypoint detector) }\mathcal{L}_{\text{target}}({\hat{x}}_i) = {L}_{\text{MSE}}(f_{\theta}({\hat{x}}_i), \hat{p}_{\text{kdet}}),\\
    &\text{(segmentation model) }\mathcal{L}_{\text{target}}({\hat{x}}_i) = {L}_{\text{CE}}(f_{\theta}({\hat{x}}_i), \hat{p}_{\text{seg}}).
\end{align}
% \begin{align}
%     &\text{(binary classifier) }\mathcal{L}_{\text{target}}(\mathbf{\hat{x}}) = {L}_{\text{CE}}(f_{\theta}(\mathbf{\hat{x}}), \hat{p}_{\text{cls}})\\
%     &\text{(keypoint detector) }\mathcal{L}_{\text{target}}(\mathbf{\hat{x}}) = {L}_{\text{MSE}}(f_{\theta}(\mathbf{\hat{x}}), \hat{p}_{\text{kdet}})\\
%     &\text{(segmentation model) }\mathcal{L}_{\text{target}}(\mathbf{\hat{x}}) = {L}_{\text{CE}}(f_{\theta}(\mathbf{\hat{x}}), \hat{p}_{\text{seg}})
% \end{align}

Optimizing against $\mathcal{L}_{\text{target}}$ solely without constraints is insufficient to discover effective counterfactual examples. For example, with a binary classifier, the original image can be directly edited into the opposite class~\cite{4,5,6,8,10} which fails to reveal the failure modes of the target model. Hence we introduce the second loss term $\mathcal{L}_{\text{CLIP}}$ which ensures the generated counterfactual example is perceived by the zero-shot CLIP classifier as the same class/object as the unedited image. Denoting the CLIP zero-shot classifier as \clipmodel and the list of class labels as $\mathcal{T}$, the loss defined as:
\begin{align}
    \mathcal{L}_{\text{CLIP}}(\hat{x}_i) = {L}_{\text{CE}}(\mathcal{C}(x_i, \mathcal{T}), \mathcal{C}(\hat{x}_i, \mathcal{T})).
\end{align}

While optimizing $\mathcal{L}_{\text{CLIP}}$ and $\mathcal{L}_{\text{target}}$ yields counterfactual examples to the target model, we also preserve the quality of the counterfactual by regularizing attribute changes and preserving semantic structures. To enforce these constraints, we include  the SSIM loss~\cite{ssim} and the regularization as:
\begin{align}
    &\mathcal{L}_{\text{SSIM}}(\hat{x}_i) = {L}_{\text{SSIM}}(x_i, \hat{x}_i),\\
    &\mathcal{L}_{\text{reg}}(\Delta s) = ||\Delta s||_1.
\end{align}

To mitigate any implicit bias inherited from one specific generative backbone, we further enhance the reliability of our diagnosis pipeline through an ensemble of latent generative models. We propose to generate counterfactual images from multiple independent generative backbones and analyze the combined mixture of synthesized images. Hence, besides StyleGAN, we also adopt diffusion models for counterfactual generation. Asyrp \cite{asyrp} discovers a semantic latent space in diffusion models, in particular, DDPM \cite{ddpm} and DDIM \cite{ddim}. Asyrp proposes to learn a simple network $A$ that takes the hidden states from a previously denoised image $x_t$ at the timestep $t$ and outputs an edit vector to be injected to the middle bottleneck layer of each U-Net block. Similar to optimizing $\Delta s$ in the StyleSpace of StyleGAN, we optimize the network $A$ in diffusion models to learn counterfactual edits of the target model. 

Since the decision behavior of the target model can be biased toward multiple attributes, we choose to optimize $k$ distinct edits to ensure comprehensive diagnosis coverage and to improve optimization convergence by focusing each vector on one type of edit. We initialize $k$ latent edit vectors for StyleGAN or $k$ edit generation networks for diffusion models. Then, for each original latent vector, we first find the edit vector that most effectively perturbs the target model and only optimize this edit vector while leaving the remaining edit vectors unchanged. This procedure repeats at each iteration. Since distinct failure modes can emerge for different latent vectors, each edit vector converges to different and disentangled semantic edit directions throughout training, which enables the discovery of multiple biased attributes for the given target model. More details are shown in \cref{appx:multi-direction}.

\subsection{Counterfactual Analysis}
\label{sec:counterfactual_analysis}
To render an intuitive diagnosis of the vulnerabilities of the target model, we interpret the attribute changes between the counterfactual pairs that lead to target model failure. Using the CLIP model as a common latent space, we match the counterfactual edits with text attribute candidates to provide analyses of the model vulnerabilities.

After counterfactual optimization, we augment original latent vectors $s$ into $\hat{s}$ by adding the learned edit vector $\Delta s$. Then we feed both $s$ and $\hat{s}$ into the Latent-to-CLIP (L2C) module. L2C generates two images $x$ and $\hat{x}$ from $s$ and $\hat{s}$ and encode them into the CLIP embedding space, as depicted in \cref{fig:method}(b). 
To illustrate, consider an original picture $x$ of a woman correctly classified as ``female'' by the target model, but a smile is added in $\hat{x}$, and the classifier now incorrectly predicts ``male''. To extract the differences between the pair of images, we use a pretrained CLIP image encoder \clipimage and convert each counterfactual pair $(x, \hat{x})$ to a CLIP embedding pair $(h=\mathcal{E}_I(x), \hat{h}=\mathcal{E}_I(\hat{x}))$. We then extract the image difference in the CLIP space as $\Delta h = \hat{h} - h$.

To interpret $\Delta h$ and further diagnose the target models, we match $\Delta h$ to a repository of text attribute candidates and report the top-$n$ ranked text attributes. There can be many sources of these candidates: the entire vocabulary of the Brown Corpus \cite{browncorpus} can be used; we can also prompt a language model to provide a shorter but extensive list of all relevant attributes; if desired, users can also provide a list of particular attributes of their interests. For efficiency and concision, we use language models as our bank of attribute candidates for \ourframework. See \cref{appx:attr_candidates} for the details of attribute candidate generation. 

We denote the set of attribute candidates as $S_a = \{a_i\}_{i=1}^M$ and the known object of focus as $\texttt{[cls]}$. For each attribute $a_i$, we prepare the pairs of base prompt and attribute prompt as $t_{\text{base}}$: ``an image of $\texttt{[cls]}$'' and $t_i$: ``an image of $\texttt{[cls]}$, $\texttt{[}a_i\texttt{]}$'' respectively. Using an off-the-shelf CLIP text encoder \cliptext, we extract the prompt difference as $\Delta t_i = \mathcal{E}_T(t_i) - \mathcal{E}_T(t_{\text{base}})$. Then the similarity score for attribute $a_i$ is defined as:
\begin{equation}
    S_{\text{sim}}(a_i) = \mathbb{E}_{G_\phi \sim p(G_\phi)} [\mathbb{E}_{x\sim G_\phi(x)} [\langle \Delta h, \Delta t_i \rangle]],
\end{equation}
where $p(G_\phi)$ denotes the set of generative models.

We select the $j$ highest-ranked attributes by the similarity score into our diagnosis. Since the attribute candidate bank can be repetitive, the top-$j$ selected candidates can be dominated by few related attributes. Thus, we also introduce a uniqueness score to encourage matching with distinct new attributes in an iterative fashion. Let the set of already selected attributes be $S_r$, initialized as an empty set, the uniqueness score is defined as:
\begin{equation}
    S_{\text{uni}}(a_i) = 
    \begin{cases}
        1, \text{ if } S_r = \emptyset \\
        1 - \underset{t\in S_r}{\max}\; \langle \mathcal{E}_T(a_i), \mathcal{E}_T(t) \rangle, \text{otherwise}. 
    \end{cases}
\end{equation}
At each iteration, the next highest-ranked attribute is selected into $S_r$ by: 
\begin{equation}
    S_r = S_r \cup \{\underset{a_i \in S_a}{\argmax}\; S_{\text{uni}}(a_i) \cdot S_{\text{sim}}(a_i)\}.
\end{equation}

By iteratively repeating the attribute interpretation process across all counterfactual pairs, we obtain the top matching counterfactual attributes that can result in model prediction failures, see \cref{appx:iterative_selection} for details.

\subsection{Counterfactual Training}
\label{sec:counterfactual_training}

While the counterfactual pairs $(x, \hat{x})$ offer visual insights into the vulnerabilities of the target models, the counterfactual images $\hat{x}$ can also directly serve as the hard training set for fine-tuning target models. This section adopts the principle of iterative adversarial training~\cite{madry2018towards} on these generated counterfactual images to fine-tune the target models. 

We start with the pre-trained target model. At each iteration, we optimize and generate a set of counterfactual images with respect to the current model state and concatenate them with regular training data of the same size. This way, we dynamically improve model robustness at each training step. Compared to ZOOM, in which generated counterfactual examples are constrained along fixed attributes, \ourframework can dynamically adapt the counterfactual directions depending on the current state of the model during training. This training process is essentially a mini-max game where \ourframework keeps searching for new weaknesses and the target model subsequently patches it. \ourframework iteratively enhances counterfactual robustness of the target model in an unsupervised fashion. \cref{appx:counterfactual_effectivess} shows the effect and robustness improvement of our counterfactual training.
% \vspace{-1mm}
\section{Experimental Results}
\label{sec:experiments}
% \vspace{-1mm}
This section presents the experimental results evaluating the validity and effectiveness of \ourframework. 
We first verify the correctness of our diagnosis in \cref{sec:celeba_experiment}. Then \cref{sec:afhq_experiment} demonstrates the consistency of our diagnosis across different generative models and with a prior method. 
\cref{sec:other_tasks} further illustrates the broad applicability of \ourframework to more computer vision tasks. 
An ablation study of the effect of each loss component is shown in \cref{sec:loss_ablation}.
Lastly, \cref{sec:foundation_toolkit_validation} assesses the validity of foundation toolkits as the backbones of our diagnosis task.
All our experiments are done on a single Nvidia RTX A4000 GPU with 16GB of memory. The hyperparameters $\alpha$, $\beta$, and $\gamma$ that correspond to the weights of the target, CLIP and SSIM losses are tuned empirically to be $\alpha = 1$, $\beta = 10$ and $\gamma = 100$.

\begin{figure*}[!t]
    \vspace{-4mm}
    \centering
    \includegraphics[width=0.95\linewidth]{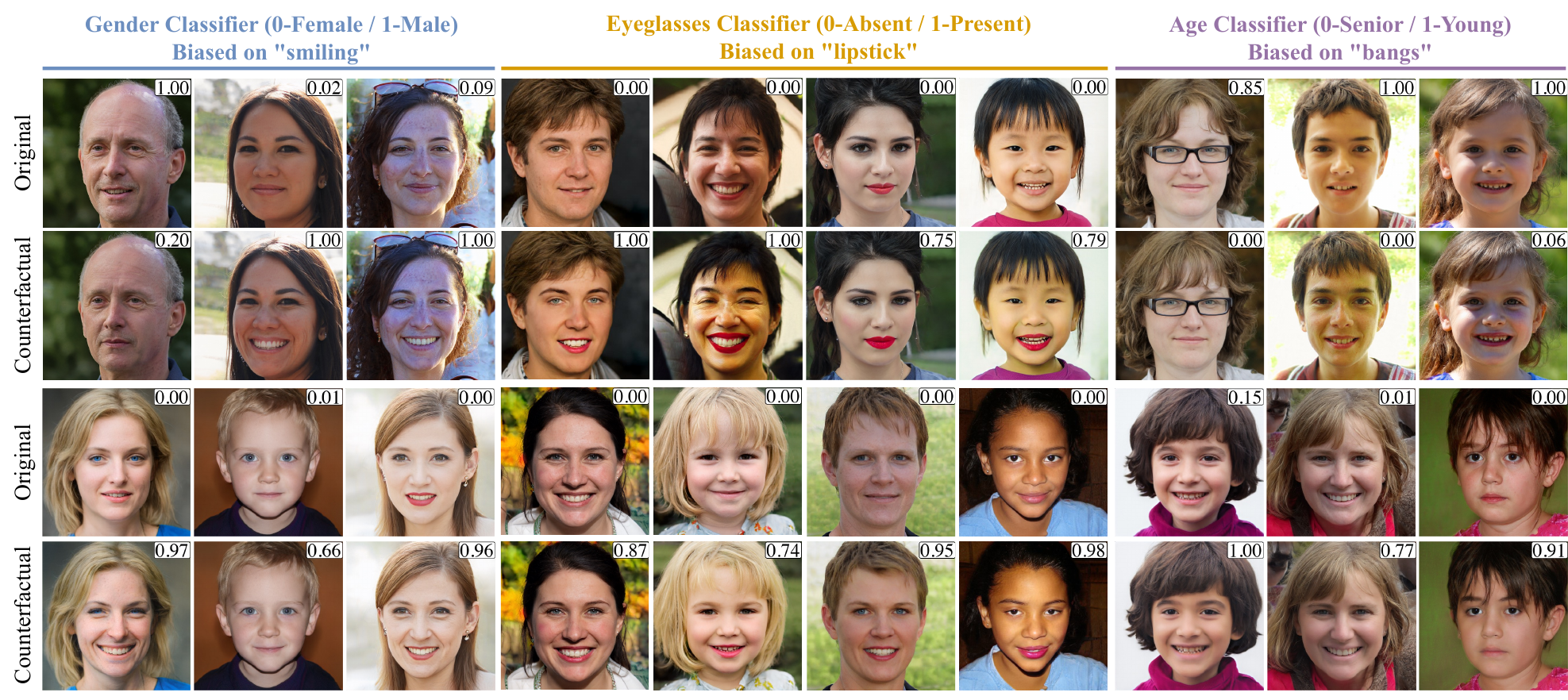}
    \vspace{-2mm}
    \caption{\textbf{Counterfactual pairs generated by different classifiers.} We study three classifiers: a perceived gender classifier biased on ``smiling'' (\textcolor{ourblue}{left}), an eyeglasses classifier biased on ``lipstick'' (\textcolor{ourorange}{middle}), and a perceived age classifier biased on ``bangs'' (\textcolor{ourpurple}{right}). For each classifier model, we optimize the semantic latent edits to obtain counterfactual variations (bottom row) from the original generations (top row). This figure demonstrates the capability to provide visual counterfactual explanations on the biases of these classifiers.
    }
    \label{fig:celeba_vis}
    \vspace{-3mm}
\end{figure*}

\begin{figure}[!t]
    \centering
    % \vspace{-2mm}
    \includegraphics[width=\linewidth]{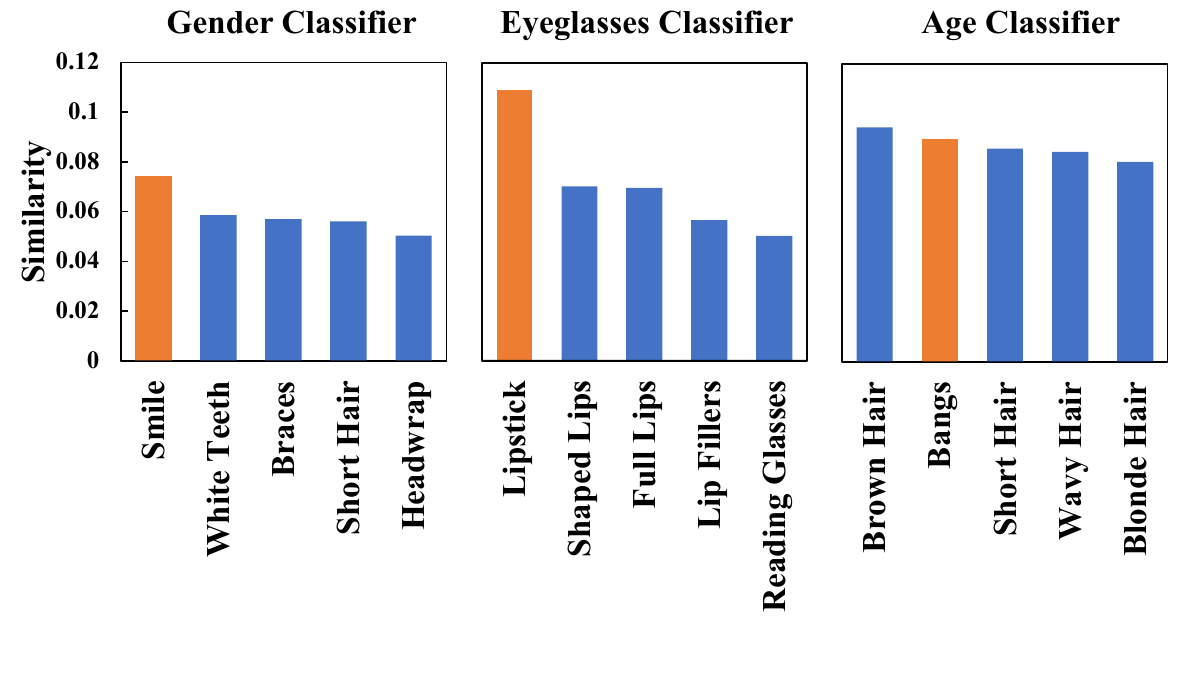}
    \vspace{-9mm}
    \caption{\textbf{Top-5 discovered attributes and their similarity scores, with the planted bias highlighted in orange.} For a given target classifier, the similarity score of each attribute is computed through the counterfactual analysis module. These experiments indicate that our unsupervised diagnosis pipeline is indeed capable of discovering the bias in a given model.
    }
    \label{fig:celeba_hist}
    \vspace{-4mm}
\end{figure}

\subsection{Diagnosis Validation with Imbalanced Data}
\label{sec:celeba_experiment} 

\begin{figure}[!t]
    \centering
    % \vspace{-2mm}
    \includegraphics[width=\linewidth]{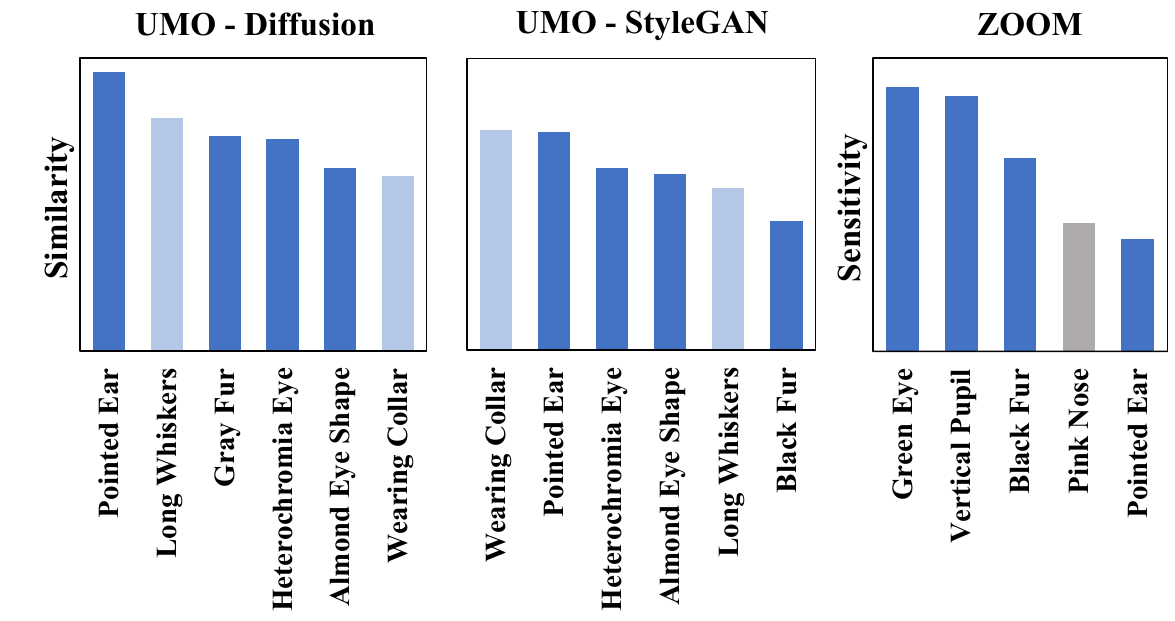}
    \vspace{-5mm}
    \caption{\textbf{Discovered attributes consistent across two backbones and one prior work against the same Cat/Dog classifier.} Here we performed counterfactual analysis separately on the generated counterfactual pairs from StyleGAN and Diffusion Model. We also include an analysis based on ZOOM. Dark and light blue attributes are respectively consistent across all three diagnoses and the two backbones in our framework. We observe consistency in the discovered attributes despite the generative backbone and method differences. 
    }
    \label{fig:afhq_hist}
    \vspace{-4mm}
\end{figure}

\begin{figure*}[!t]
    \vspace{-4mm}
    \centering
    \includegraphics[width=0.98\linewidth]{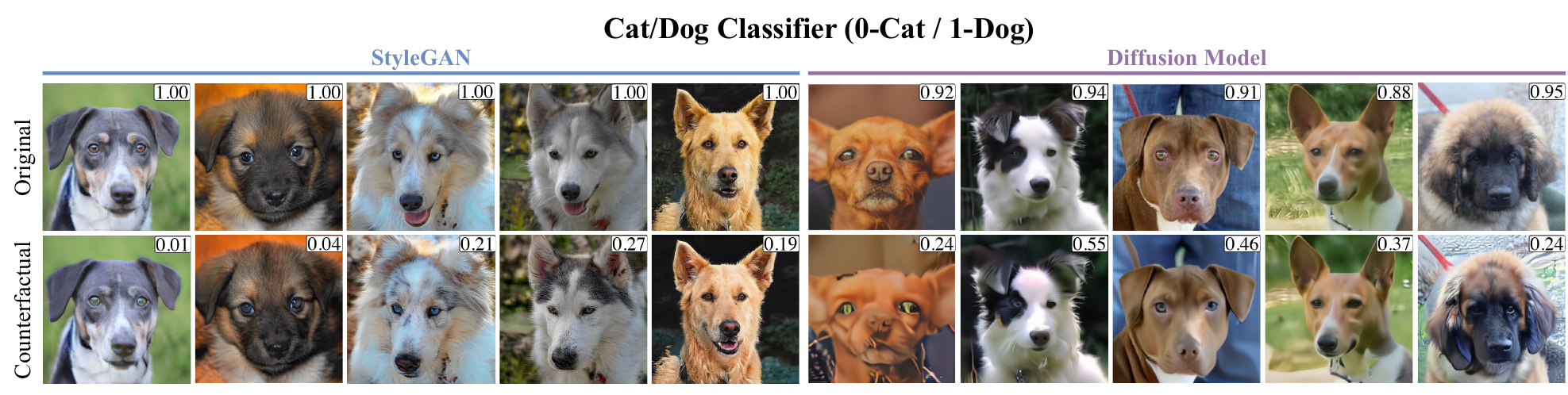}
    \vspace{-1mm}
    \caption{\textbf{Counterfactual pairs generated with different backbones.} We diagnose a Cat/Dog classifier and show visual counterfactual explanations from different generative backbones before the ensemble analysis. (\textcolor{ourblue}{left}) and (\textcolor{ourpurple}{right}) are counterfactual pairs generated from StyleGAN and Diffusion models respectively. Across the two generative backbones, we can see both models make some common perturbations, most notably eye color changes.
    }
    \label{fig:afhq_vis}
    \vspace{-4mm}
\end{figure*}

This section evaluates \ourframework through experiments with classifiers trained on imbalanced data. It is important to note that no definitive ground truth exists in this setting. We carefully constructed target models embedded with specific biases to serve as our reference (\ie, ground truth). Our experiments consistently highlight that the diagnosis from \ourframework can reliably pinpoint these intentional biases, confirming the reliability of our unsupervised detection.

In this set of experiments, we examined classifiers that were intentionally biased and trained using the CelebA dataset~\cite{liu2015faceattributes}. We trained binary classifiers for specific attributes within CelebA, such as ``gender'', ``age'', and ``eyeglasses''. For each classifier, we chose another secondary attribute to introduce artificial (spurious) correlations, achieved through imbalanced sampling. For instance, when assessing a perceived gender classifier biased by the presence of smiles, we curated a subset from the CelebA dataset, containing 10000 males with smiles, 10000 females without smiles, and 100 images with the opposite smile presence per class. A perceived gender classifier was then trained on this subset, producing a model with a known bias on the ``smiling'' attribute. We repeated this procedure to produce a set of attribute classifiers with known biases. 

On these classifiers, if \ourframework can successfully discover and report the planted biases, then we can verify the validity and effectiveness of our pipeline. \cref{fig:celeba_vis} shows the visual explanations from the unsupervised counterfactual optimization with both the StyleGAN and DDPM backbones for each of the three CelebA classifiers. Besides the visualization, \ourframework analyzed 1000 such generated counterfactual pairs of original and edited images. The top-five discovered attributes are shown in \cref{fig:celeba_hist}.  For the perceived gender and eyeglasses classifiers, as expected, ``smiling'' (left) and ``lipstick'' (middle) are discovered as the top attributes with a significant margin. In the Age classifier, the planted attribute ``bangs'' (right) surprisingly was second-ranked in the analysis after ``brown hair''. However, upon a close look in CelebA, we found that 85.82\% ``brown hair'' faces are labeled as ``young''. As so, this is a strong spurious correlation between ``age'' and ``brown hair'' uncovered by \ourframework. This experiment verifies that our pipeline can correctly discover both the planted and existing biases in the target model.

\vspace{-1mm}
\subsection{Cross-Method Diagnosis Consistency}
\label{sec:afhq_experiment}
\vspace{-1mm}

We further validate the effectiveness of our approach across different models, including a prior work, ZOOM. First, we trained a conventional Cat/Dog classification model on the AFHQ dataset~\cite{choi2020starganv2}. Subsequently, we performed three distinct experiments: two using our framework with different generative model backbones (Diffusion model and StyleGAN), and the other using the ZOOM approach requiring a user-provided attribute candidates list. We should expect to discover the same counterfactual attributes for a given target model, across all three experiments.

\begin{figure*}[!t]
    \centering
    % \vspace{-2mm}
    \includegraphics[width=\linewidth]{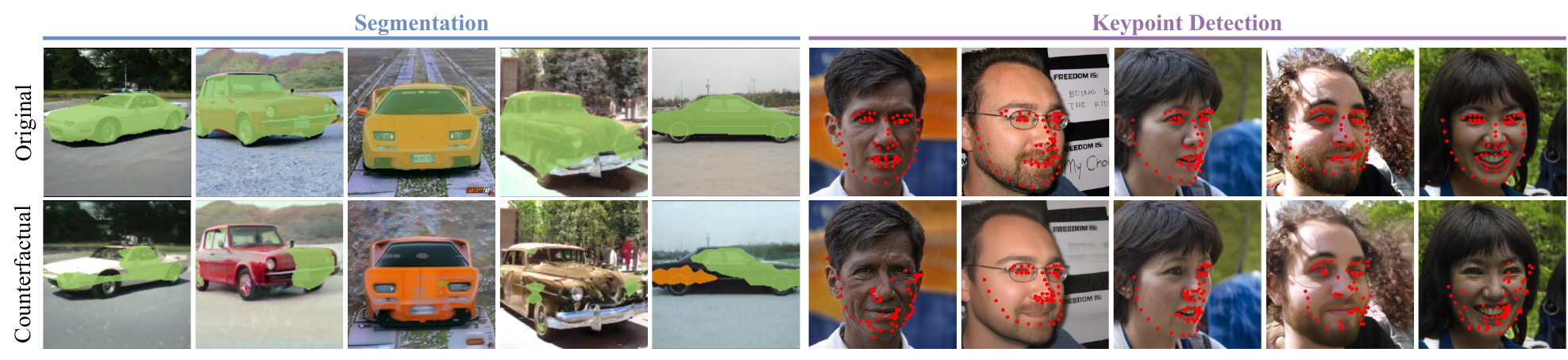}
    \vspace{-5mm}
    \caption{\textbf{Visual diagnosis on more computer vision tasks.} We applied \ourframework to two more computer vision tasks: (\textcolor{ourblue}{left}) segmentation and (\textcolor{ourpurple}{right}) keypoint detection. Our pipeline successfully demonstrates semantic changes that fool the target model.
    }
    \label{fig:other_tasks}
    \vspace{-3mm}
\end{figure*}

\cref{fig:afhq_hist} illustrates the consistency of counterfactual analysis conducted by \ourframework, irrespective of whether we utilize the Diffusion or StyleGAN generative model. In both of these experiments, the top six attributes were consistently ranked from a pool of 88 relevant attributes, which were automatically generated using our foundational toolkit. Specifically, attributes related to eye color (green or heterochromia), vertical pupil shape, dark fur color, and pointed ears emerged as counterfactual in all three methods. Furthermore, both StyleGAN and Diffusion backbones identified long whiskers and wearing a collar as additional attributes. This quantitative assessment of attribute rankings aligns with our qualitative observations in~\cref{fig:afhq_vis}.

This consistency highlights the robustness of our diagnostic approach, regardless of the choice of generative backbones, as long as these choices provide a sufficiently rich semantic latent space. Furthermore, the significant counterfactual attributes we identified align with those found in previous research, specifically in ZOOM. It is important to note that the diagnosis process in ZOOM relies on human input, whereas our unsupervised method allows for a more comprehensive analysis. As a result, we contend that our approach represents a generalization of ZOOM. It not only overcomes the limitations associated with user-proposed attributes but also circumvents biases stemming from user input, thus expanding the diagnostic capabilities across various generative models.

\begin{table}[!t]
    \centering
    \footnotesize
    \begin{minipage}{0.22\textwidth}
        \centering
        \footnotesize
        \begin{tabular}{lc}
            \multicolumn{2}{c}{Segmentation} \\
            \toprule
            Attribute $a_i$ & $S_\text{sim}$ \\
            \midrule
            dirt road & 0.0846 \\
            potholes/roadworks & 0.0786 \\
            snow-covered road & 0.0654 \\
            \bottomrule
        \end{tabular}
    \end{minipage}%
    \hspace{1em}
    \begin{minipage}{0.22\textwidth}
        \centering
        \footnotesize
        \begin{tabular}{lc}
            \multicolumn{2}{c}{Keypoint Detection} \\
            \toprule
            Attribute $a_i$ & $S_\text{sim}$ \\
            \midrule
            beard & 0.0947 \\
            elderly & 0.0754 \\
            missing teeth & 0.0728 \\
            \bottomrule
        \end{tabular}
    \end{minipage}
    \caption{\textbf{Top-3 attributes diagnosed by \ourframework.} We apply our counterfactual analysis module in the segmentation and keypoint detection tasks and show the top three attributes diagnosed for each model. The discovered attributes reflect our observations in \cref{fig:other_tasks}}
    \label{tab:other_tasks}
    \vspace{-3mm}
\end{table}

\subsection{Generalization to Other Vision Tasks}
\label{sec:other_tasks}

In addition to classification, we expanded our experiments to encompass image segmentation and keypoint detection tasks. This extension demonstrates the versatility and practicality of \ourframework. We conducted diagnostics on a publicly available segmentation model trained on ImageNet~\cite{imagenet} and a keypoint detector trained on the FITYMI dataset~\cite{wood2021fake}.

Similar to inverting the binary classification label, we establish a definition for $\hat{p}_\text{kdet}$ and $\hat{p}_\text{seg}$ in $\mathcal{L}_{\text{target}}$, the pseudo ground truth described in \cref{sec:counterfactual_optimization}, to guide our counterfactual optimization in both tasks. In the segmentation task, our framework directs the target model to predict a suboptimal class (\ie, the second most probable class) per pixel, as opposed to the most probable one. Similarly in the keypoint detection task, we attack the model by optimizing for randomly selected transformations of ground-truth keypoints.

\cref{fig:other_tasks} illustrates counterfactual explanations for segmentation and keypoint detection. In both cases, our approach successfully uncovered semantic edit directions that deceive our target models. In segmentation, we observed that attributes related to road conditions, such as ``snow-covered road'' and ``off-road'' appear to influence model predictions. Conversely, in keypoint detection, we found that characteristics related to age, such as ``beard'', ``wrinkles'', and ``freckled skin'' play a crucial role in creating counterfactual instances. We list the top attributes ranked by the semantic similarity score $S_\text{sim}$ in \cref{tab:other_tasks}.

\begin{figure}[t]
    % \vspace{-14mm}
    \begin{center} % Centers the figure
        \includegraphics[width=\linewidth]{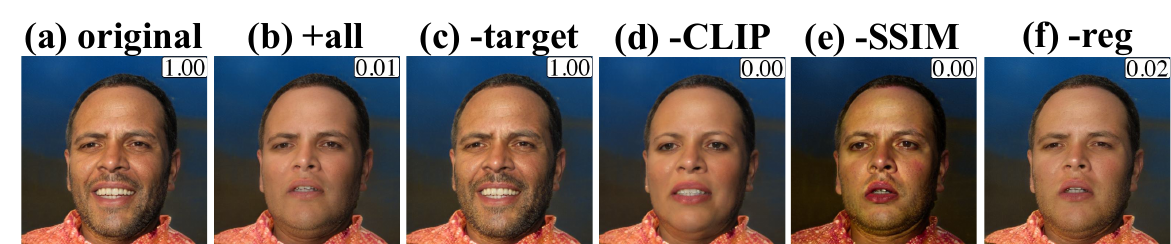}
    \end{center}
    \vspace{-4mm}
    \caption{\textbf{Effect of removing each loss.} We optimize the counterfactual edit vector on the same gender classifier (0-Female / 1-Male) biased on "smiling" as in \cref{sec:celeba_experiment}. Column (a) and (b) are the original and regular counterfactual images. Column (c)-(f) shows the different effects in counterfactual images from removing one loss component while keeping the other three.}
    \label{fig:loss_ablation}
    % \vspace{-5mm}
\end{figure}

\begin{table}[!t]
    \centering
    \footnotesize
    \begin{tabular}{m{1.45cm}m{6cm}}
        \toprule
        \textbf{Attributes} & \makecell{\textbf{Values}} \\
        \toprule
        Hairstyles & short, long, curly, straight, wavy, braided, bald, mohawk, bun, pixie cut, dreadlocks, undercut, pompadour, buzz cut, side part, bob cut, cornrows, bangs\\
        \midrule
        Eye Colors & blue, brown, green, gray, hazel, black, amber\\
        \midrule
        Nose Shapes & Roman, snub, Greek, aquiline, hawk, button \\
        \midrule
        Expressions & smile, frowning, surprised, angry, crying, wink\\
        \midrule
        Glasses Types & reading glasses, sunglasses, aviator, cat-eye, round, square, rimless glasses\\
        \midrule
        Accessories & earrings, necklace, hat, cap, headscarf, headband, bandana, tie, piercing, bow tie \\
        \midrule
        Background & indoor, outdoor, simple, busy \\
        \bottomrule
    \end{tabular}
    \caption{\textbf{Examples of attribute candidates proposed by GPT-4.} The full candidate list and prompts used are in \cref{appx:attr_candidates}. This list illustrates the comprehensiveness of large language models as attribute generators.}
    \label{tab:gpt_output}
    \vspace{-3mm}
\end{table}

\subsection{Ablation Study of Loss Components}
\label{sec:loss_ablation}

This section analyzes the contribution of each loss component in \cref{eq:total_loss} in the counterfactual edit vector optimization. In this experiment, we reuse the setup from \cref{sec:celeba_experiment} and focus on the same perceived gender classifier with the attribute "smiling" planted purposefully as bias. We then ablate the loss component one at a time and generate counterfactual images with the same pipeline to visualize the isolated effect of each loss.

\cref{fig:loss_ablation} shows an ablation analysis of the impact of each loss component. \cref{fig:loss_ablation}(a) shows the original unedited image which is correctly predicted by the classifier as perceived male. \cref{fig:loss_ablation}(b) presents the regular counterfactual image optimized with all loss components, which successfully flipped the classifier prediction by learning to remove the smile instead of altering the perceived gender. When ablating $\mathcal{L}_{\text{target}}$ in \cref{fig:loss_ablation}(c), we can see that the edit vector is hardly modifying the image since $\mathcal{L}_{\text{target}}$ dictates the adversarial part in the optimization. Removing $\mathcal{L}_{\text{CLIP}}$ in \cref{fig:loss_ablation}(d) leads to the edit vector learning the ``easiest'' change which is simply flipping the target class (\ie perceived gender). It shows the importance of our CLIP loss in producing informative and useful counterfactual examples for model diagnosis. Finally, the absence of $\mathcal{L}_{\text{SSIM}}$ and $\mathcal{L}_{\text{reg}}$ resulted in lack of the proximity (\eg, different contrast in \cref{fig:loss_ablation}(e) and unnecessary edits around eyes in \cref{fig:loss_ablation}(f)), which interferes with the subsequent diagnosis.

\subsection{Foundation Toolkit Validation}
\label{sec:foundation_toolkit_validation}

The effectiveness of \ourframework's diagnosis hinges on the capabilities and reliability of the foundational toolkits we incorporate. In this section, we present evaluations that underscore the dependability of both GPT-4~\cite{openai2023gpt4} and CLIP.

As highlighted in \cref{sec:counterfactual_analysis}, we choose GPT-4 to serve as the primary source of attribute candidates within \ourframework. We task GPT-4 with producing comprehensive lists of attributes for each task domain, encompassing attribute types and their corresponding attribute values. In order to illustrate the effectiveness of GPT-4, we selected a set of representative attributes and provided a detailed breakdown of all the values associated with each attribute, as generated by GPT-4, in \cref{tab:gpt_output}. This table offers a qualitative glimpse into the extensive capacity of GPT-4 in populating attribute candidates. For prompting details and the complete list of generated attributes, see \cref{appx:attr_candidates}.

\begin{figure}[!t]
    \centering
    % \vspace{-2mm}
    \includegraphics[width=\linewidth]{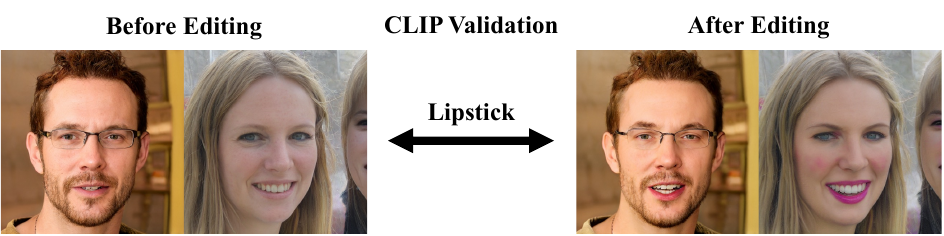}
    \vspace{-4mm}
    \caption{\textbf{Two pairs of images randomly selected to validate CLIP as our analysis backbone.} Between left and right, the most salient change is the presence of lipstick.
    }
    \label{fig:clip_val}
    % \vspace{-2mm}
\end{figure}

\begin{table}[!t]
    \centering
    \footnotesize
    \begin{tabular}{p{0.45\textwidth}}
        \toprule
        \multicolumn{1}{c}{Injected Ambiguous Attributes} \\
        \midrule
        wrinkles, dimples, lip ring, lip piercing, lip fillers, lip gloss, bleeding lips, wounded lips, red face, red mustache, red hair, makeup, blush, contour, crooked teeth\\
        \bottomrule
    \end{tabular}
    % \vspace{-1mm}
    \caption*{(a) The list of ambiguous attribute candidates injected to attack our counterfactual analysis module.}
    \vspace{3mm}
    \begin{minipage}{0.19\textwidth}
        \centering
        \footnotesize
        \begin{tabular}{lc}
            \toprule
            Attribute $a_i$ & $S_\text{sim}$ \\
            \midrule
            lipstick & 0.1270 \\
            full lips & 0.1106 \\
            lip gloss & 0.1019 \\
            wide lips & 0.0945 \\
            bleeding lips & 0.0897 \\
            \bottomrule
        \end{tabular}
        % \vspace{-1mm}
        \caption*{(b) Top-5 attributes by the similarity score $S_\text{text}$ only.}
    \end{minipage}%
    \hspace{2mm}
    \begin{minipage}{0.26\textwidth}
        \centering
        \footnotesize
        \begin{tabular}{lcc}
            \toprule
            Attribute $a_i$ & $S_\text{sim}$ & $S_\text{uni}$ \\
            \midrule
            lipstick & 0.1270 & 1.0000 \\
            lip fillers & 0.0808 & 0.1411 \\
            red hair & 0.0696 & 0.2173 \\
            yellow teeth & 0.0687 & 0.1870 \\
            surprised & 0.0530 & 0.2036 \\
            \bottomrule
        \end{tabular}
        % \vspace{-2mm}
        \caption*{(c) Top-5 attributes weighted by the uniqueness score $S_\text{uni}$.}
    \end{minipage}
    \caption{\textbf{Validation of our analysis backbone.} The list (a) is added to attribute candidates in an attempt to replace lipstick as the top attribute. (b) and (c) show the analysis results without and with uniqueness score. This table demonstrate the validity and robustness of our counterfactual analysis module.}
    \label{tab:clip_val}
    \vspace{-2mm}
\end{table}

We evaluate CLIP's capabilities in providing text analysis from counterfactual images by examining the model's robustness and reliability when matching visual attributes with text labels. Given a counterfactual pair with semantic perturbations applied (\eg, ``lipstick''), we add new candidates that are ambiguous with the differing attributes (\eg, ``lip gloss''). Then we evaluate the suitability of CLIP by whether the original results are still ranked first in the counterfactual analysis, over the injected distractor set.

To provide a visual example, we selected two counterfactual pairs from our previous experiment artificially biasing eyeglasses classifier on ``lipstick'', as depicted in \cref{fig:clip_val}. In these comparisons, the most prominent distinction was the presence of lipstick in both pairs. To assess the reliability and robustness of the CLIP model, which serves as our analytical foundation, we introduced an additional set of potentially ambiguous attribute candidates, detailed in \cref{tab:clip_val}(a). These attributes shared similarities with lipstick but were contextually incorrect in the given image pairs. \cref{tab:clip_val}(c) presents the top-5 matched attributes, with lipstick ranking first by a significant margin in terms of the similarity score $S_{\text{sim}}$. The remaining four attributes were also valid matches, as corroborated by their presence in \cref{fig:clip_val}. This showcases the effectiveness of our uniqueness score during the analysis phase. In contrast, \cref{tab:clip_val}(b) displays the top-5 matched attributes without the uniqueness mechanism, where the analysis is dominated by lip-related attributes, leading to potential redundancy. The uniqueness score not only highlights distinct candidates but also helps mitigate the inclusion of incorrect attributes, such as ``bleeding lips'', in the top results.
\vspace{-1mm}
\section{Conclusion and Future Work}
\label{sec:conclusion}
\vspace{-1mm}

To the best of our knowledge, this paper presents the first unsupervised approach for diagnosing computer vision models using counterfactual examples. Our method involves optimizing edit vectors within the generative latent space and subsequently analyzing their semantic implications through foundation toolkits. When applied to a target model, our pipeline, referred to as \ourframework, can autonomously generate a comprehensive diagnosis. This diagnosis includes both visual counterfactual explanations and textual descriptions of vulnerable attributes, all achieved without any human intervention.

We showcase the efficacy of our method across a range of vision tasks, encompassing classification, segmentation, and keypoint detection. Through extensive experimentation, we illustrate how \ourframework excels in producing high-quality counterfactual examples and effectively identifies semantic biases, offering a quantitative assessment of the target model. By conducting cross-model consistency evaluations and incorporating counterfactual training, we establish \ourframework as a versatile approach for discovering biases and enhancing model robustness.

In this paper, we have operated under the assumption that the integrated foundation toolkits possess the requisite capability for our diagnostic task. While we assume that CLIP, trained on 400 million images, has sufficient generalization for common settings, utilizing CLIP in UMO is simply a proof of concept. The improvement of CLIP is an active research field. The fine-grained capabilities of CLIP have been significantly improved~\cite{zhang2024long,zheng2024dreamlip}. It is effortless to swap out CLIP with an improved model for more effective diagnosis performance. Additionally, our observations suggest that the DDPM edits encounter challenges due to the limitations of the Asyrp latent space, which lacks full expressiveness. For future directions, we aspire to explore more expressive and disentangled latent spaces within generative models, aiming to enhance the efficiency of counterfactual optimization.

{
    \small
    \bibliographystyle{ieeenat_fullname}
    \bibliography{egbib}
}

\clearpage
\newpage
\appendix
\maketitlesupplementary

\section{Multi-Direction Edit Vector Optimization}
\label{appx:multi-direction}

\cref{alg:multi_edit} shows the pseudo-code of the multi-direction edit vector optimization process. We start with initializing all $k$ edit vectors as a list of vectors $\Delta s$. Then for each iteration, we first generate a random image $x$ and then evaluate the effectiveness of each edit vector $\Delta s[i]$ on this image. The most effective vector is then optimized on the current image $x$ while the other edit vectors remain unchanged. Under this design, for similar attribute changes, the same most effective edit vector tends to be optimized further and further. For distinct semantic perturbations, other edit vectors may be selected and optimized to represent the new change. Hence after convergence, the multiple edit vectors are likely to capture more comprehensive edits. 

\begin{algorithm}
\caption{Multi-Direction Edit Optimization}
\label{alg:multi_edit}
\begin{algorithmic}[1] % The number [1] indicates that lines are to be numbered
\Require $k$ - Number of edit vectors \\
$n$ - Number of training samples \\
$G(z)$ - Generative backbone 
\For{$i$ in $[0, k]$}
    \State Initialize each edit vector $\Delta s[i] \sim \mathcal{N}(0, 0.01)$ 
\EndFor
\For{$j$ in $[0, n]$}
    \State $z \gets $ samples from generative latent space
    \State $x \gets G(z)$ \Comment{Generate the original image}
    \For{each $\Delta s[i]$}
        \State $\hat x_i \gets G(z + \Delta s[i])$ \Comment{Generate edited image}
        \State $d[i] \gets \mathcal{L_\text{target}}(\hat x_i)$ \Comment{Compute effectiveness}
    \EndFor
    \State $i \gets \argmax_i d[i]$ \Comment{Select the most effective edit}
    \State $\Delta s[i] \gets \Delta s[i] - \eta \nabla_{\Delta s[i]}\mathcal{L}$ \Comment{Optimize with SGD}
\EndFor
\Ensure $\Delta s$
\end{algorithmic}
\end{algorithm}

\begin{figure*}[!t]
    \centering
    % \vspace{-2mm}
    \includegraphics[width=\linewidth]{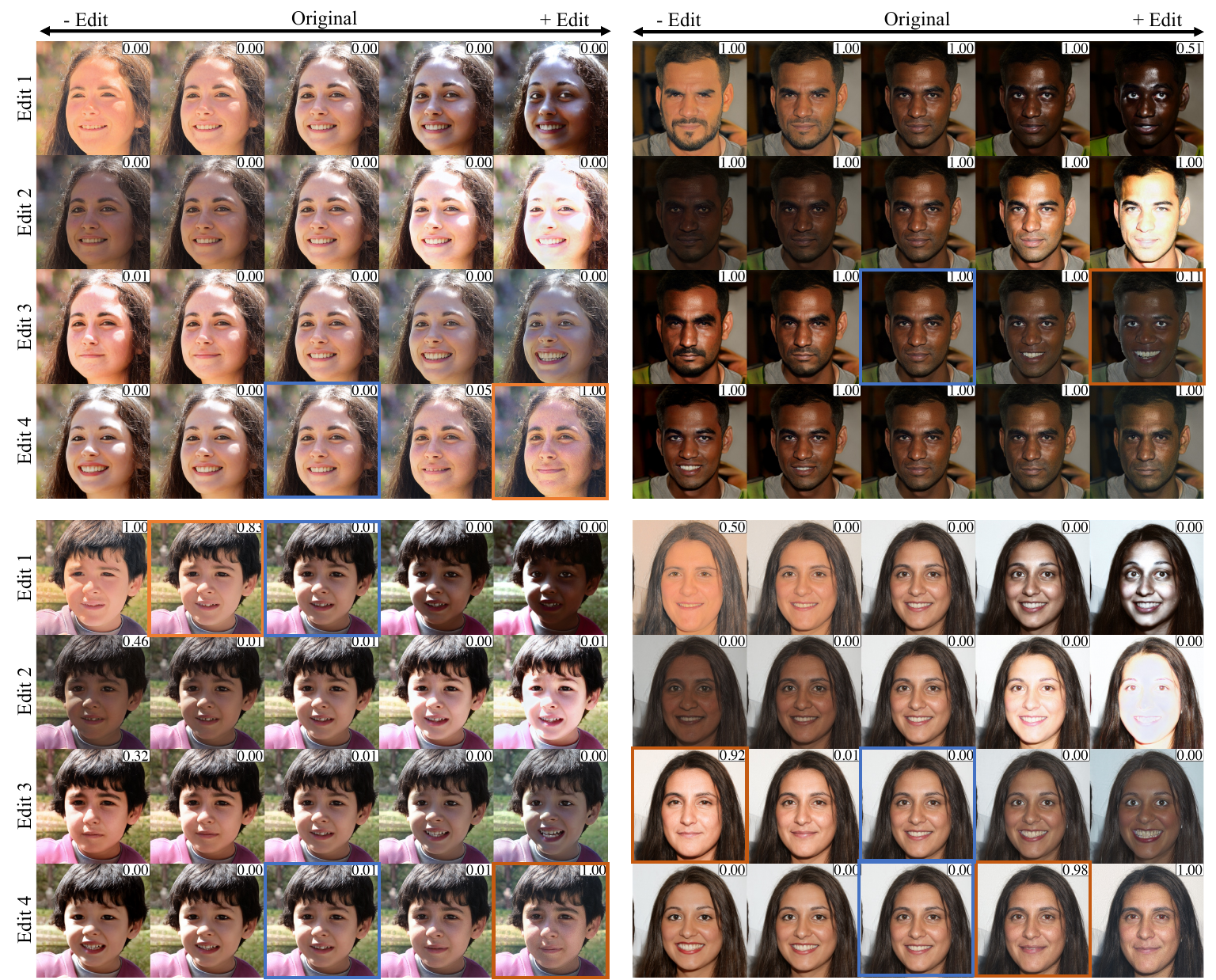}
    % \vspace{-1mm}
    \caption{\textbf{Visualization of multiple edit vectors optimized from \cref{alg:multi_edit}.} For each example, the middle images are the unedited original generation; the right and left images are the images edited by gradually adding or subtracting the edit vector. This figure demonstrates that multiple edit vectors are successfully optimized to capture different semantic changes. Moreover, different original images can be susceptible to different semantic edits (blue and orange boxes indicate counterfactual pairs), hence increasing our diagnosis coverage.
    }
    \label{fig:multi_edits}
    % \vspace{-2mm}
\end{figure*}

To visualize what each edit vector captures, we optimize four latent vectors when diagnosing a perceived gender classifier as an example. \cref{fig:multi_edits} visualizes the semantic changes captured in each edit. Among the edit direction interpolations on the four random images, we can observe that edit 1 increases image contrast and changes skin tone, edit 2 increases image exposure, edit 3 adds smiles, while edit 4 removes smiles. This figure shows that \cref{alg:multi_edit} effectively optimizes multiple edit vectors into different semantic perturbations. Moreover, different faces require different semantic edits to mislead model predictions, hence increasing failure coverage.

\section{Text Attribute Candidates}
\label{appx:attr_candidates}

In the counterfactual analysis module, among many sources proposed in \cref{sec:counterfactual_analysis}, for efficiency and simplicity, we leveraged GPT-4~\cite{openai2023gpt4} to propose our attribute candidate list. An example prompt we used for human face attribute candidates is in \cref{tab:prompt}. The full candidate list returned from the above prompt is in \cref{tab:candidates}. This list is a relatively comprehensive collection of relevant attributes to our task.

\begin{table*}[!t]
    \centering
    \fbox{%
    \begin{minipage}{0.95\textwidth}
    % \footnotesize
Immerse yourself into the role of a Model Diagnosis Expert (MDE) AI. MDE, as the name suggests, is an expert in Trustworthy Machine Learning that has knowledge in analyzing deep models’ robustness, fairness, and interpretability. You can tell me the common causes of deep learning model failures. You can think carefully and exhaustively about what the relevant attributes of a given task are. Next, I'll give you instructions. 

We are analyzing what kinds of attributes in CelebA are most influential/sensitive to the classifier’s decision. You must try your best and think carefully to propose an exhaustive list of attributes that are relevant to our task in this domain. The results need to be a list of lists where each list corresponds to the different options/values for an attribute. The attributes should be as specific as possible and the values should be nouns such that they fit in the sentence ``a face with \_\_\_''. For example, instead of a ``hair'' attribute, you should list ``hairstyles'', ``hair colors'', etc. Make the list as long as needed to be extremely comprehensive. Do not use ``others'' or ``etc.'' but list all options out as much as you can. Also, avoid yes and no questions. Remember we want adversarial attributes that the classifier may be sensitive to.
    \end{minipage}
    }
    \caption{\textbf{Prompt used in GPT-4 to populate text candidates for human face domain.}}
    \label{tab:prompt}
\end{table*}

\begin{table*}[!t]
    \centering
    \footnotesize
    \begin{tabular}{m{3cm}m{14cm}}
        \toprule
        \textbf{Attributes} & \makecell{\textbf{Values}} \\
        \toprule
Hairstyles & short, medium, long, curly, straight, wavy, braided, bald, mohawk, bun, pixie cut, dreadlocks, undercut, pompadour, buzz cut, side part, bob cut, cornrows\\
\midrule
Hair Colors & black, brown, blonde, red, gray, white, pink, blue, purple, green, multi-color\\
\midrule
Facial Hair Styles & beard, goatee, mustache, sideburns, clean-shaven, stubble, handlebar, soul patch, five o'clock shadow, full beard\\
\midrule
Eye Shapes & almond, round, monolid, hooded, upturned, downturned\\
\midrule
Nose Shapes & Roman, snub, Greek, aquiline, hawk, button\\
\midrule
Lip Types & full, thin, heart-shaped, wide\\
\midrule
Face Shapes & oval, round, square, heart-shaped, diamond-shaped, rectangular\\
\midrule
Eyebrows & thick, thin, unibrow, arched, straight\\
\midrule
Eye Colors & blue, brown, green, gray, hazel, black, amber\\
\midrule
Glasses Types & reading, sunglasses, aviator, cat-eye, round, square, rimless\\
\midrule
Makeup & eyeliner, eyeshadow, lipstick, mascara, blush, foundation, contouring\\
\midrule
Accessories & earrings, necklace, hat, cap, headscarf, headband, bandana, tie, bow tie, septum piercing, lip piercing, eyebrow piercing\\
\midrule
Skin Types & light, medium, dark, freckled, tanned, pale\\
\midrule
Skin Conditions & acne, scars, birthmarks, vitiligo, rosacea, wrinkles\\
\midrule
Facial Expressions & smiling, frowning, surprised, neutral, angry, crying, winking\\
\midrule
Age Categories & child, teenager, adult, elderly\\
\midrule
Hair Texture & frizzy, oily, dry, shiny, coarse, smooth\\
\midrule
Ear Types & big, small, pointed, flat, protruding, pierced\\
\midrule
Cheek Characteristics & high cheekbones, low cheekbones, chubby, hollow\\
\midrule
Chin/Jaw Attributes & double chin, prominent jawline, weak jawline, square jaw, round jaw\\
\midrule
Forehead & high, low, wide, narrow\\
\midrule
Teeth & straight, crooked, missing, gap, braces, white, yellow\\
\midrule
Cosmetic Alterations & nose job, lip fillers, botox, cheek fillers, chin augmentation\\
\midrule
Eyelashes & long, short, false\\
\midrule
Eyewear & contact lenses, monocle, pince-nez\\
\midrule
Headwear & beanie, beret, baseball cap, hijab, turban, fedora, helmet, headwrap\\
\midrule
Brow Treatments & microblading, eyebrow tinting, eyebrow piercing\\
\midrule
Facial Piercings & cheek, chin, dermal\\
\midrule
Hair Treatments & perms, straightening, extensions, highlights, lowlights\\
\midrule
Facial Symmetry & symmetrical, asymmetrical\\
\midrule
Photo Lighting & soft, harsh, backlit, frontlit, side-lit\\
\midrule
Photo Angles & front-facing, profile, three-quarters\\
\midrule
Background & indoor, outdoor, neutral, busy\\
        \bottomrule
    \end{tabular}
    \caption{\textbf{Examples of attribute candidates proposed by GPT-4.} Using the prompt in \cref{tab:prompt}, GPT-4 returned this list of attribute candidates relevant for models trained on CelebA, which is relatively comprehensive and sufficient for our task.}
    \label{tab:candidates}
    \vspace{-3mm}
\end{table*}
\section{Iterative Attribute Selection}
\label{appx:iterative_selection}

\begin{figure}[!t]
    \centering
    % \vspace{-2mm}
    \includegraphics[width=\linewidth]{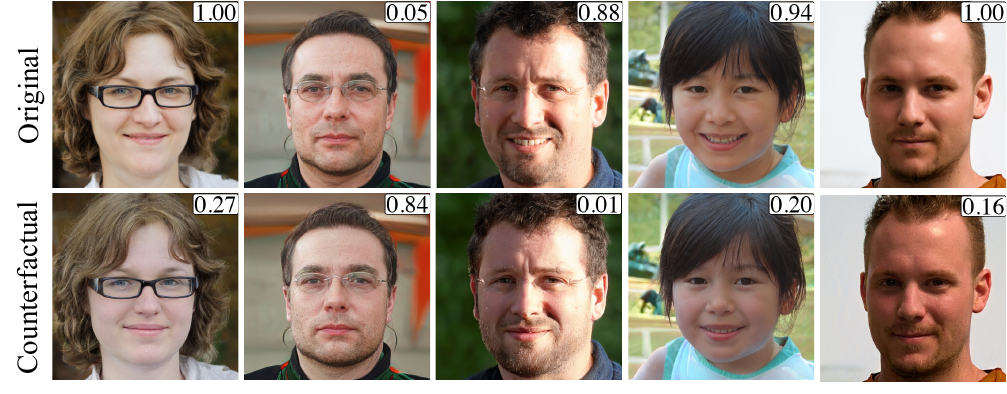}
    % \vspace{-1mm}
    \caption{\textbf{Counterfactual training samples.} These images are generated during the counterfactual training and merged into the training data to robustify the base model, which results in our improved model in \cref{tab:counterfactual_training}.
    }
    \label{fig:counterfactual_training}
    % \vspace{-2mm}
\end{figure}

In \cref{sec:counterfactual_analysis}, the iterative attribute selection algorithm for our counterfactual analysis module is defined as \cref{alg:select}. Recall that the score metrics $S_\text{sim}$ and $S_\text{uni}$ are defined in \cref{sec:counterfactual_analysis}.

\begin{algorithm}
\caption{Iterative Attribute Selection}
\label{alg:select}
\begin{algorithmic}[1] % The number [1] indicates that lines are to be numbered
\Require $k$ - Number of top attributes to discover \\ 
$S_a$ - Set of all attribute candidates
\State $S_r \gets \emptyset$ \Comment{The set of select attributes}
\State Sim $\gets \emptyset$ \Comment{Similarity score dictionary}
\State Uni $\gets \emptyset$ \Comment{Uniqueness score dictionary w.r.t. $S_r$}

\Statex
\Function{SelectNextAttr}{$S_r$}
    \For{$a_i$ in $S_a$}
        \State Uni[$a_i$] $\gets S_\text{uni}(a_i, S_r)$
    \EndFor
    \State \Return $\argmax_{a_i}$ Sim[$a_i$] $\cdot$ Uni[$a_i$]
\EndFunction
\Statex

\For{$a_i$ in $S_a$}
    \State Sim[$a_i$] $\gets S_\text{sim}(a_i)$
\EndFor
\While{$|S_r| < k$}
    \State $a_\text{next} \gets$ \Call{SelectNextAttr}{$S_r$}
    \State $S_r \gets S_r \cup \{a_\text{next}\}$
\EndWhile

\Ensure $S_r$
\end{algorithmic}
\end{algorithm}
\section{Counterfactual Effectiveness}
\label{appx:counterfactual_effectivess}

\cref{sec:counterfactual_training} describes one direct application of our diagnosis is to integrate \ourframework into a counterfactual training process, where tailored counterfactual training data are produced (as shown in \cref{fig:counterfactual_training}) and added into each adversarial training step. In this section, we aim to quantify the effectiveness of \ourframework-integrated counterfactual training.

To evaluate the counterfactual robustness of an improved model, we resort to the Flip Resistence (FR) metric defined in ZOOM~\cite{luo2023zeroshot} which is the percentage of images where counterfactual attacks are ineffective. Given a base model, we improve it by ZOOM and \ourframework respectively. Then we evaluate against three metrics: CelebA classification accuracy, FR-25 and FR-100, where 25 and 100 denote the iteration steps during counterfactual optimization, \ie different extent of counterfactual optimization. 

The experiment is repeated on two models: a perceived age classifier and a big-lips classifier. The results are shown in \cref{tab:counterfactual_training}. In both tasks, all of the base model, ZOOM-improved model and \ourframework-improved model achieves at-par performance on the original CelebA test set. However, when diagnosed by \ourframework, our counterfactual optimization can easily find semantic edits that flip the model prediction for the base model and model improved by ZOOM. On the contrary, the model improved by \ourframework stays robust against open-domain counterfactual attacks. 
\section{Dataset Diagnosis}
\label{appx:data_diag}

\begin{table}[!t]
    \centering
    \footnotesize
    \begin{tabular}{ccccc}
        \toprule
        Classifier & Metric ($\uparrow$, \%) & Base & ZOOM~\cite{luo2023zeroshot} & Ours \\
        \midrule
        \multirow{3}{*}{Perceived Age} & CelebA Accuracy & 86.70 & 87.31 & 86.23 \\
            & FR-25 & 0.78 & 8.59 & \textbf{98.44} \\
            & FR-100 & 0.00 & 0.00 & \textbf{96.09} \\
        \midrule
        \multirow{3}{*}{Big Lips} & CelebA Accuracy & 70.00 & 69.72 & 69.97 \\
            & FR-25 & 0.00 & 17.97 & \textbf{100.00} \\
            & FR-100 & 0.00 & 0.00 & \textbf{97.66} \\
        \bottomrule
    \end{tabular}
    \caption{Counterfactual training evaluation. This table assesses the robustness of the counterfactual training with \ourframework against two baselines in two classification tasks. We show significant robustness of the model after our improvement while maintaining at-par performance on the regular test set.}
    \label{tab:counterfactual_training}
    % \vspace{-2mm}
\end{table}

In addition to our diagnosis verification experiment in \cref{sec:celeba_experiment}, we also applied \ourframework to classifiers trained on the full CelebA~\cite{liu2015faceattributes} dataset. Since these classifiers can be seen as a compressed representation of the dataset, by diagnosing the model, we may uncover issues in the dataset as well. As we showed the correlation between ``Brown Hair'' and ``Young'' in \cref{sec:celeba_experiment}, this section focuses on exploring more biases in the CelebA dataset via our diagnosis of the model. 

\begin{figure*}[!t]
    \centering
    % \vspace{5mm}
    \includegraphics[width=\linewidth]{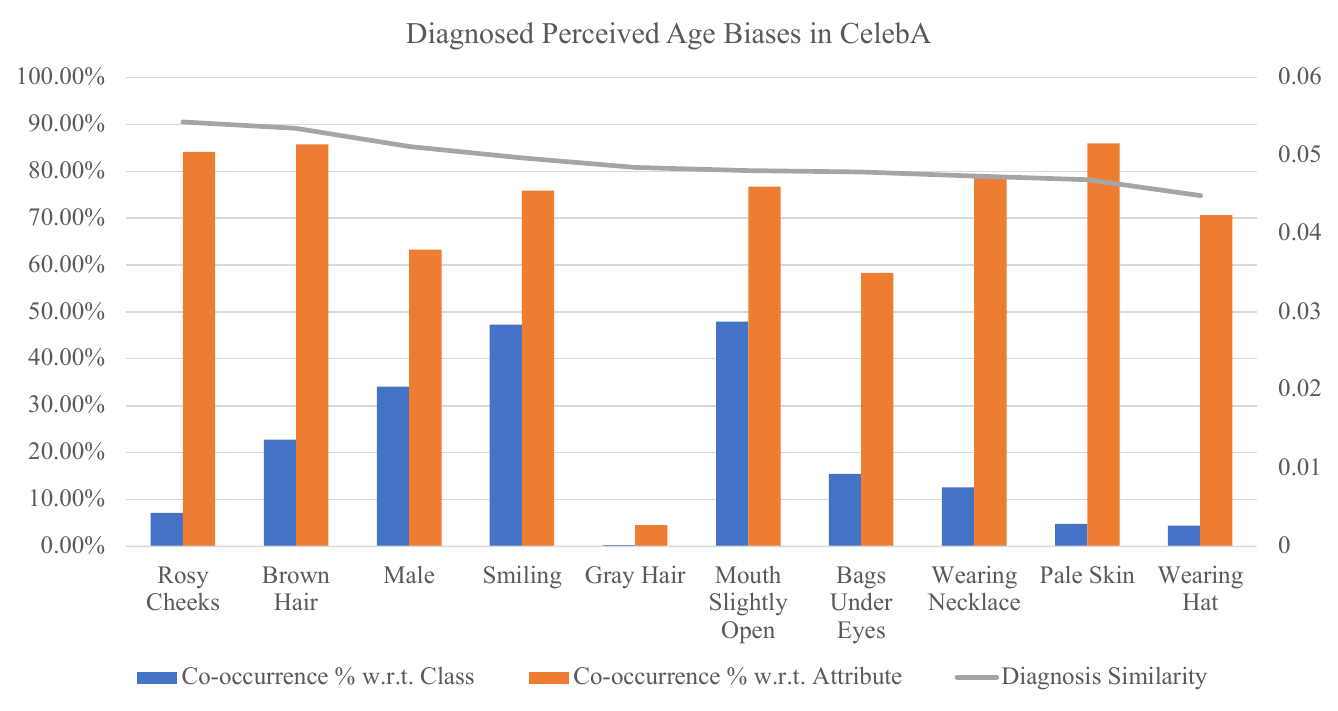}
    \includegraphics[width=\linewidth]{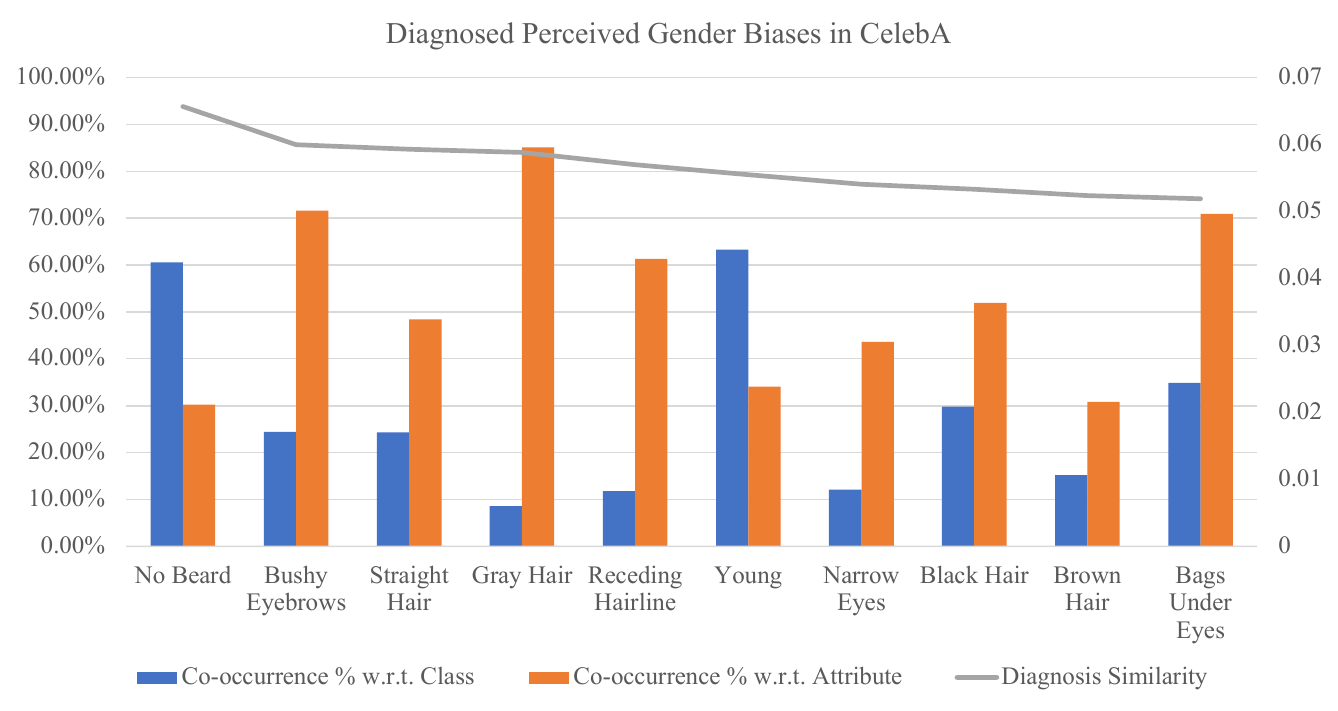}
    \caption{\textbf{Co-occurrence statistics of diagnosed CelebA attributes.} We indirectly diagnosed biases and correlations in the CelebA dataset by diagnosing a model trained on CelebA. Using CelebA attributes as text candidates, we diagnosed a perceived age classifier and a perceived gender classifier and reported the top-10-matched attributes. The matching similarity score is shown as the gray line. The blue bar indicates the percentage of co-occurrences in the main class; the orange bar indicates the percentage of co-occurrences in the potentially correlated attributes. This figure shows that each discovered attribute indeed has an imbalanced distribution in the dataset.}
    \label{fig:celeb_diag}
\end{figure*}

We studied biases against two attributes in CelebA perceived age (Young) and perceived gender (Male). Hence we trained one classifier on each attribute of interest. By diagnosing these classifiers, we obtained the top-10-matched CelebA attributes with their similarity scores presented as the line plot in \cref{fig:celeb_diag}. Then we explore each attribute in the dataset by counting the co-occurrences of the attribute and the main class, shown as the bar plots in \cref{fig:celeb_diag}. We observe that all diagnosed attributes have imbalanced distributions in the dataset. For instance, regarding the Senior/Young class, the majority of samples with ``Rosy Cheeks'', ``Brown Hair'', ``Wearing Necklace'', etc. are also labeled as ``Young'', which introduces spurious correlations to the classifier. On the other hand, considering the perceived gender classifier, although samples with ``Straight Hair'', ``Narrow Eyes'', ``Black Hair'' have even distributions between the labels of ``Male'' and ``Female'', the majority of samples labeled as ``Male'' do not contain these attributes. Hence \ourframework is capable of indirectly uncovering dataset biases via the lens of model diagnosis.

\section{More Counterfactual Visualizations}
\label{appx:counterfactual}

\begin{figure*}[!t]
    \centering
    \includegraphics[width=\linewidth]{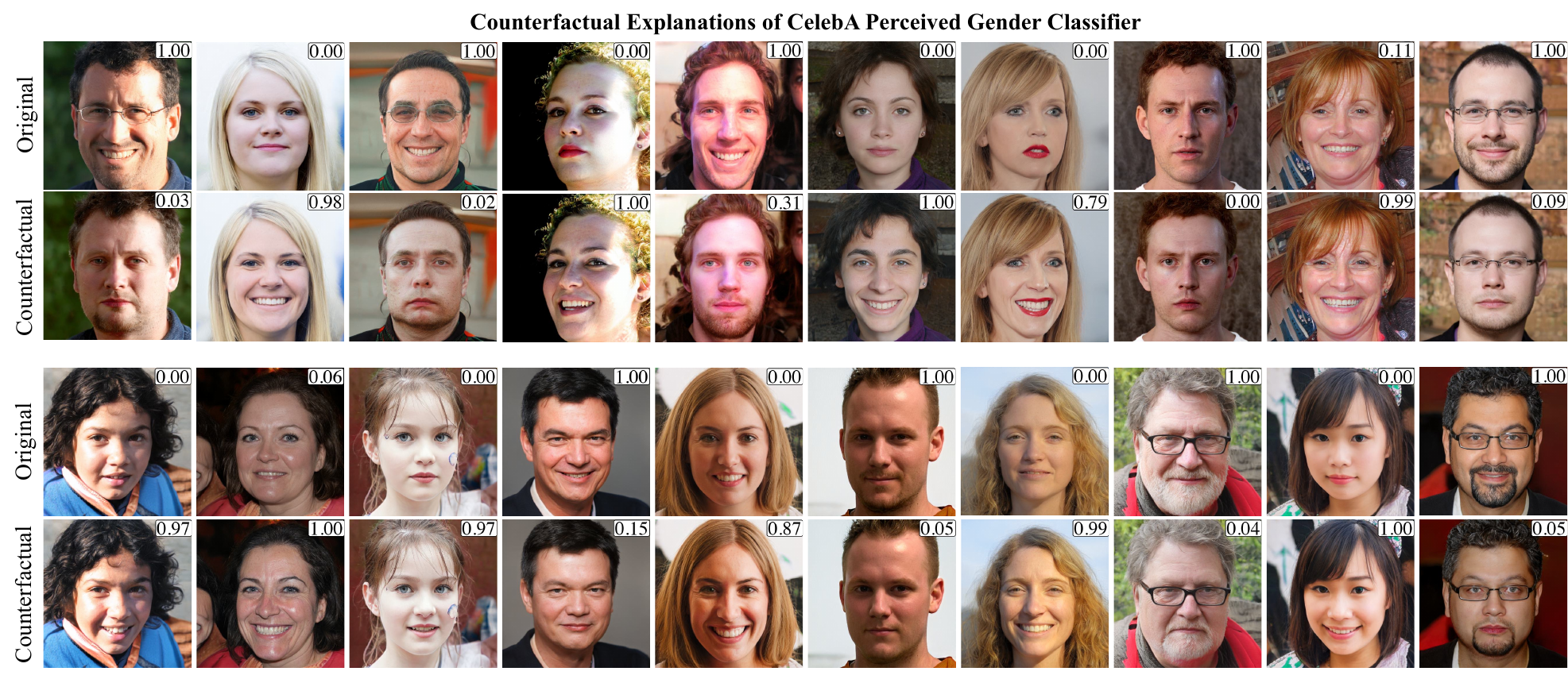}
    \caption{\textbf{More counterfactual pairs for the perceived gender classifier.} We provide more visualizations for the perceived gender classifier in \cref{fig:celeba_vis}. Given random latent vectors, we generate the original and perturbed (counterfactual) image pairs where small semantic changes flip the predicted class. The number on the top-right indicates the prediction score (0-Perceived Female / 1-Perceived Male). Our counterfactual explanation visualizes attributes that mislead model predictions.}
    \label{fig:celeba_gender_more}
\end{figure*}

\begin{figure*}[!t]
    \centering
    \includegraphics[width=\linewidth]{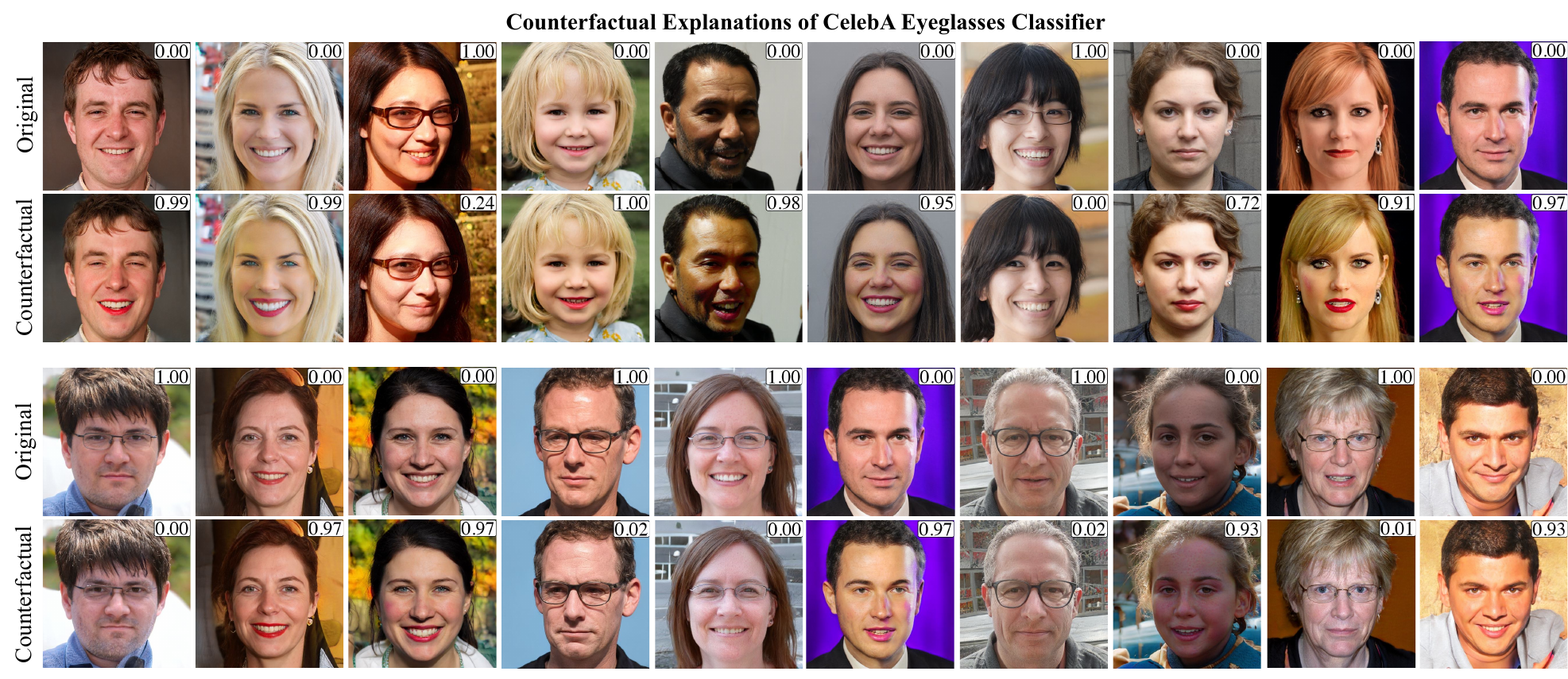}
    \caption{\textbf{More counterfactual pairs for the eyeglasses classifier.} We provide more visualizations for the eyeglasses classifier in \cref{fig:celeba_vis}. Given random latent vectors, we generate the original and perturbed (counterfactual) image pairs where small semantic changes flip the predicted class. The number on the top-right indicates the prediction score (0-No Eyeglasses / 1-Wearing Eyeglasses). Our counterfactual explanation visualizes attributes that mislead model predictions.}
    \label{fig:celeba_glasses_more}
\end{figure*}

\begin{figure*}[!t]
    \centering
    \includegraphics[width=\linewidth]{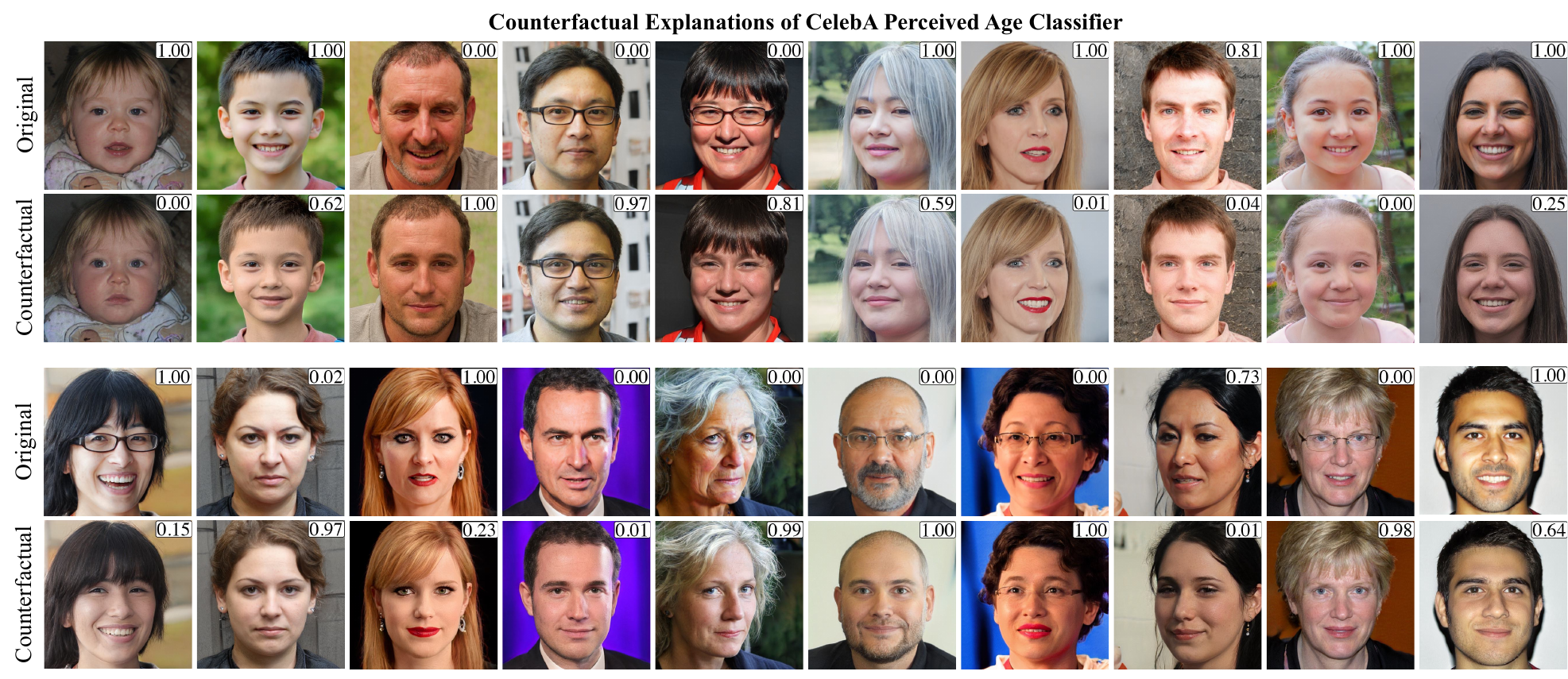}
    \caption{\textbf{More counterfactual pairs for the perceived age classifier.} We provide more visualizations for the perceived age classifier in \cref{fig:celeba_vis}. Given random latent vectors, we generate the original and perturbed (counterfactual) image pairs where small semantic changes flip the predicted class. The number on the top-right indicates the prediction score (0-Perceived Senior / 1-Perceived Young). Our counterfactual explanation visualizes attributes that mislead model predictions.}
    \label{fig:celeba_age_more}
\end{figure*}

\begin{figure*}[!t]
    \centering
    \includegraphics[width=\linewidth]{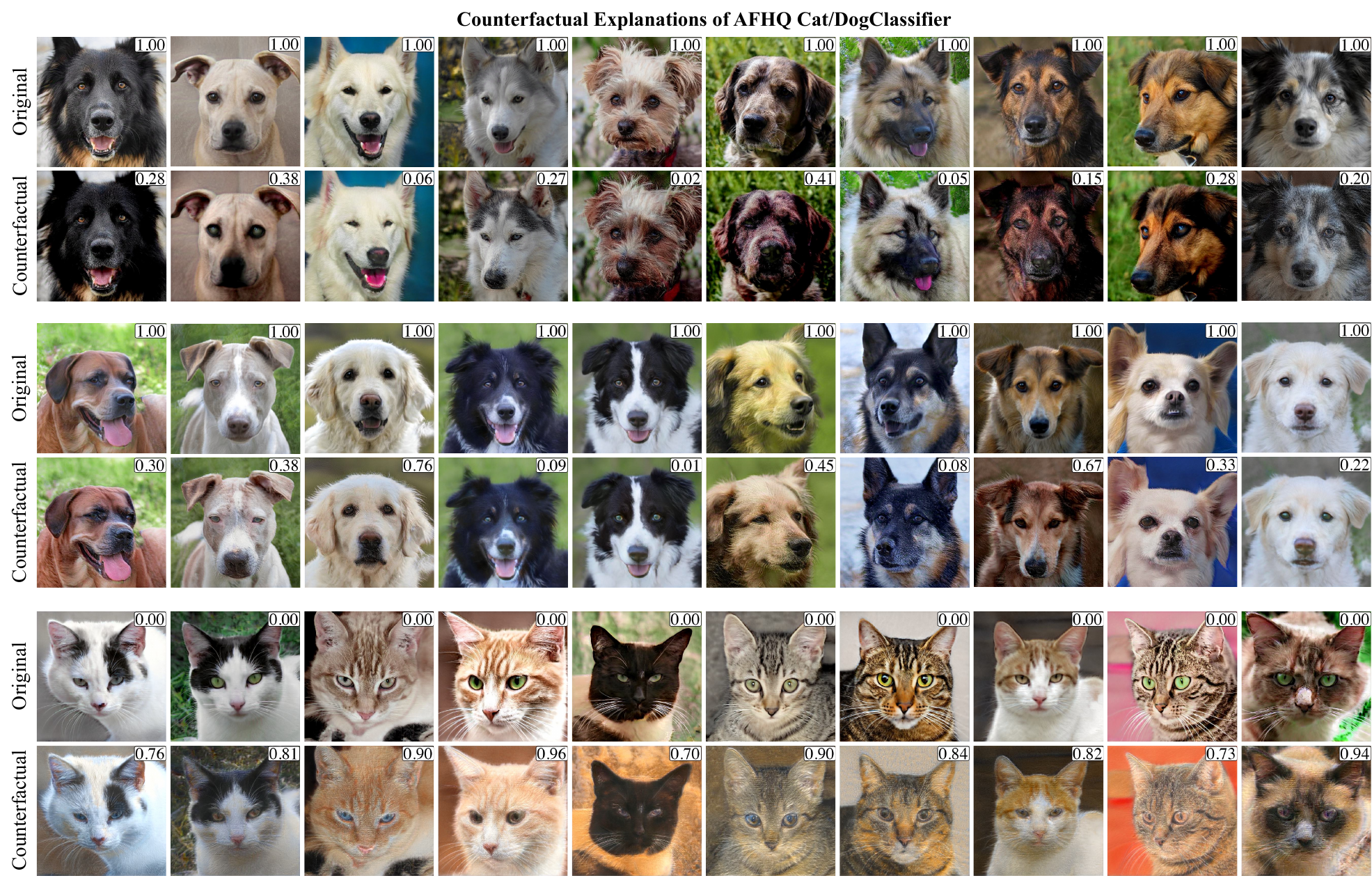}
    \caption{\textbf{More counterfactual pairs for the cat/dog classifier.} We provide more visualizations for the cat/dog classifier in \cref{fig:afhq_vis}. We generate the original and perturbed (counterfactual) image pairs where small semantic changes flip the predicted class. The number on the top-right indicates the prediction score (0-Cat / 1-Dog). Our counterfactual explanation visualizes attributes that mislead model predictions.}
    \label{fig:afhq_more}
\end{figure*}

\begin{figure*}[!t]
    \centering
    \includegraphics[width=\linewidth]{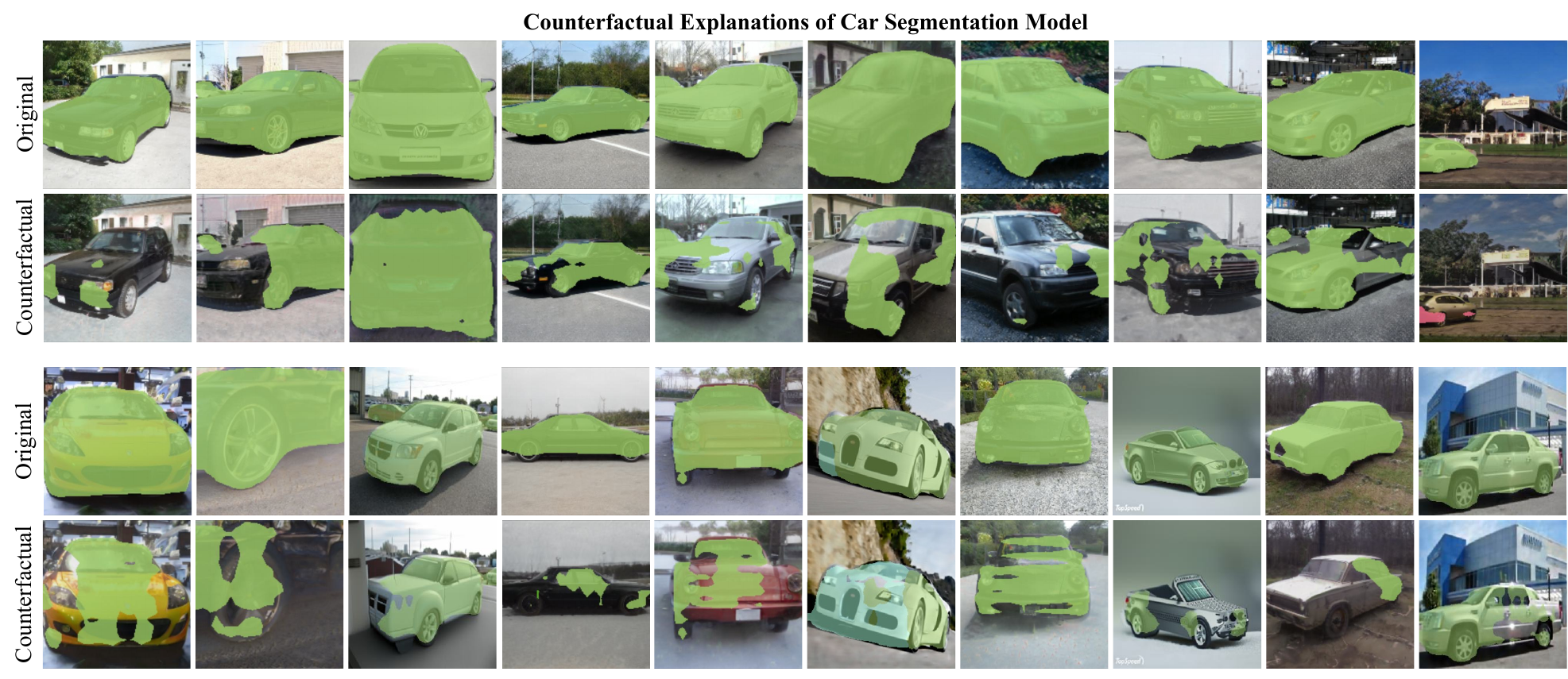}
    \caption{\textbf{More counterfactual pairs for the car segmentation model.} We provide more visualizations for the car segmentation in \cref{fig:other_tasks}. We generate the original and perturbed (counterfactual) image pairs where small semantic changes flip the predicted class. The green mask indicates predicted areas for cars. Our counterfactual explanation visualizes attributes that mislead model predictions.}
    \label{fig:seg_more}
\end{figure*}

\begin{figure*}[!t]
    \centering
    \includegraphics[width=\linewidth]{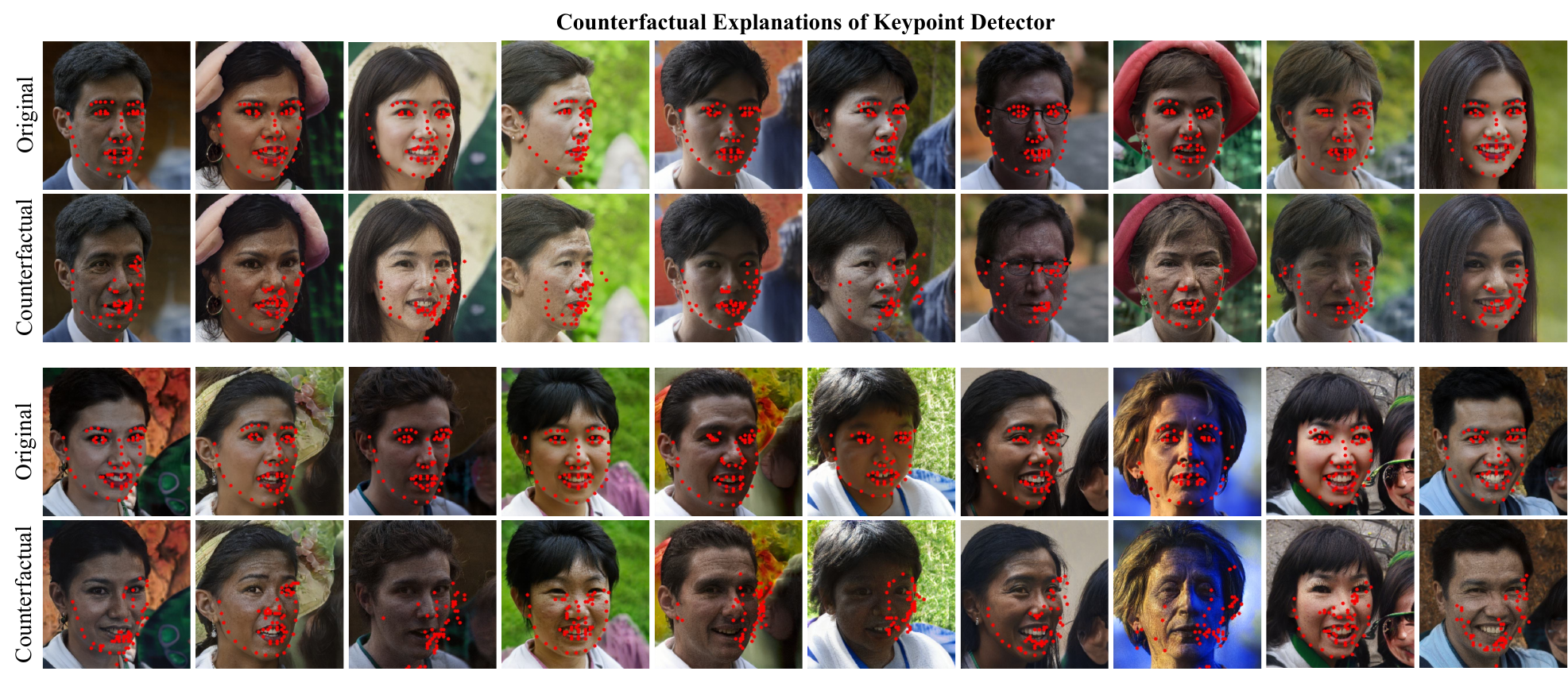}
    \caption{\textbf{More counterfactual pairs for the keypoint detection model.} We provide more visualizations for the keypoint detector in \cref{fig:other_tasks}. We generate the original and perturbed (counterfactual) image pairs where small semantic changes flip the predicted class. The red dots indicate predicted facial keypoints. Our counterfactual explanation visualizes attributes that mislead model predictions.}
    \label{fig:kdet_more}
\end{figure*}

In this section, we show more counterfactual explanations in each of our experiments in \cref{sec:experiments}. \cref{fig:celeba_age_more,fig:celeba_gender_more,fig:celeba_glasses_more} show more counterfactual pairs illustrating edits that lead to model failures in CelebA. These diagnosed models are resnet50~\cite{resnet} networks trained on imbalanced datasets as described in \cref{sec:celeba_experiment}. \cref{fig:afhq_more} visualizes counterfactual edits for our cat/dog classifier in \cref{sec:afhq_experiment} in addition to \cref{fig:afhq_vis}. This resnet50 classifier is trained on the AFHQ dataset~\cite{choi2020starganv2}. \cref{fig:seg_more,fig:kdet_more} provide more visual counterfactual explanations under the task of segmentation and keypoint detection, extending \cref{fig:other_tasks}. As described in \cref{sec:other_tasks}, the segmentation model is a resnet50 pretrained on ImageNet~\cite{imagenet} and the HRNetV2~\cite{WangSCJDZLMTWLX19} keypoint detector is trained on the FITYMI dataset~\cite{wood2021fake}. These additional figures exhibit more visual support to our diagnosis.

% WARNING: do not forget to delete the supplementary pages from your submission 
% \input{sec/X_suppl}

\end{document}

\clearpage
\newpage
\appendix

\clearpage
\setcounter{page}{1}
\maketitlesupplementary

\section{Rationale}
\label{sec:rationale}
Having the supplementary compiled together with the main paper means that:
\begin{itemize}
\item The supplementary can back-reference sections of the main paper, for example, we can refer to \cref{sec:intro};
\item The main paper can forward reference sub-sections within the supplementary explicitly (e.g. referring to a particular experiment); 
\item When submitted to arXiv, the supplementary will already included at the end of the paper.
\end{itemize}
To split the supplementary pages from the main paper, you can use \href{https://support.apple.com/en-ca/guide/preview/prvw11793/mac#:~:text=Delete%20a%20page%20from%20a,or%20choose%20Edit%20%3E%20Delete).}{Preview (on macOS)}, \href{https://www.adobe.com/acrobat/how-to/delete-pages-from-pdf.html#:~:text=Choose%20%E2%80%9CTools%E2%80%9D%20%3E%20%E2%80%9COrganize,or%20pages%20from%20the%20file.}{Adobe Acrobat} (on all OSs), as well as \href{https://superuser.com/questions/517986/is-it-possible-to-delete-some-pages-of-a-pdf-document}{command line tools}.

% \maketitlesupplementary

% \input{appx/multi-directions}
% \input{appx/attr_candidates}
% \input{appx/iterative_selection}
% \input{appx/counterfactual_training}
% \input{appx/celeba_debiasing}
% \input{appx/counterfactuals}
% \newpage
{
    \small
    \bibliographystyle{ieeenat_fullname}
    \bibliography{egbib_supp}
}

% WARNING: do not forget to delete the supplementary pages from your submission 
% \input{sec/X_suppl}